\documentclass[conference]{IEEEtran}
\IEEEoverridecommandlockouts

\usepackage{cite}
\usepackage{amsmath,amssymb,amsfonts}
\usepackage{algorithmic}
\usepackage{graphicx}
\usepackage{subcaption}  
\usepackage{caption}     
\usepackage{float}       
\usepackage{textcomp}
\usepackage{xcolor}
\usepackage{multirow}
\usepackage{graphicx}\graphicspath{{figures/}}
\usepackage{booktabs}     
\usepackage{array}        
\usepackage{caption}      
\usepackage{siunitx}      
\usepackage{paralist}

\usepackage{hyperref}

\usepackage{flushend}

\newcommand{\sstitle}[1]{\smallskip\noindent\textbf{#1.} }

\def\Snospace~{\S{}}

\newcommand\Mark[1]{\textsuperscript#1}
\def\ua{\Mark{*}}

\begin{document}

\setlength{\belowdisplayskip}{1pt}
\setlength{\belowdisplayshortskip}{1pt}
\setlength{\abovedisplayskip}{1pt}
\setlength{\abovedisplayshortskip}{1pt}

\title{An Efficient and Effective Evaluator for Text2SQL Models on Unseen and Unlabeled Data}

\author{Trinh Pham\Mark{1}, 
Thanh Tam Nguyen\Mark{1},
Viet Huynh\Mark{2},
Hongzhi Yin\Mark{3}\ua\thanks{\ua Corresponding authors}, 
Quoc Viet Hung Nguyen\Mark{1}\ua
\vspace{1.6mm}\\
\fontsize{9}{9}\selectfont\rmfamily\itshape
\small 
\Mark{1}Griffith University (Australia),
\Mark{2}Edith Cowan University (Australia),
\Mark{3}The University of Queensland (Australia)
}


\maketitle

\begin{abstract}
Recent advances in large language models have strengthened Text2SQL systems that translate natural language questions into database queries. A persistent deployment challenge is to assess a newly trained Text2SQL system on an unseen and unlabeled dataset when no verified answers are available. This situation arises frequently because database content and structure evolve, privacy policies slow manual review, and carefully written SQL labels are costly and time-consuming. Without timely evaluation, organizations cannot approve releases or detect failures early. FusionSQL addresses this gap by working with any Text2SQL models and estimating accuracy without reference labels, allowing teams to measure quality on unseen and unlabeled datasets. It analyzes patterns in the system's own outputs to characterize how the target dataset differs from the material used during training. FusionSQL supports pre-release checks, continuous monitoring of new databases, and detection of quality decline. Experiments across diverse application settings and question types show that FusionSQL closely follows actual accuracy and reliably signals emerging issues. Our code is available at \url{https://github.com/phkhanhtrinh23/FusionSQL}.
\end{abstract}


\section{Introduction}
\label{sec:introduction}

Text2SQL models translate natural-language questions into executable database queries, enabling non-experts to access relational data for analytics, business intelligence, governance, and operations \cite{li2024dawn}. The first generation of neural parsers showed that models could learn this mapping by using both the question and the database schema \cite{zhong2017seq2sql,xu2017sqlnet,yu2018typesql,yu-etal-2018-syntaxsqlnet}. Building on this base, later studies improved how models connect question words to tables and columns, and how they handle more complex query structures \cite{zhang2019editing,guo2019towards,wang2019rat,lin2020bridging}. Subsequent work introduced stronger decoding and parsing strategies, which further improved query generation \cite{rubin2020smbop,cao2021lgesql}. Large-scale pretraining then helped models transfer better across different databases \cite{yu2020grappa}. More recently, large language models have driven another leap forward through constrained generation, task decomposition, and self-correction \cite{scholak2021picard,pourreza2023din,gao2023text,talaei2024chess,LuoXY25,Le_Pham_Quan_Luu_2024,pham-etal-2024-unibridge}. To make systems more robust, recent methods also use modular designs, candidate selection, and agent-based reasoning to improve reliability in complex settings \cite{li2023resdsql,pourreza2024chase,sheng2025csc,liu2025xiyan,FanRHHW25,PanZZXZSBCCZ25,LiCSGY25,sun2025agenticdata,deng2025reforce}.

\begin{figure}[t]
    \centering
    \includegraphics[width=\linewidth]{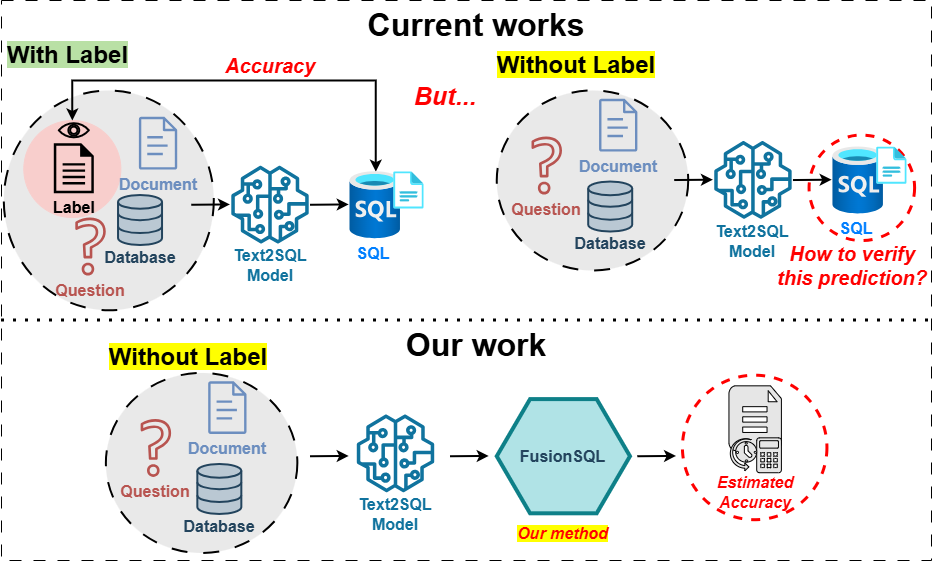}
    \caption{\textbf{Top}: Existing Text2SQL evaluations rely on ground-truth labels, which are often unavailable as databases evolve. \textbf{Bottom}: FusionSQL estimates model accuracy directly from unlabeled inputs without requiring ground-truth SQL labels.}
    \vspace{-1em}
    \label{fig:intro-pipeline}
    \vspace{-1em}
\end{figure}

Despite the rapid progress of Text2SQL models, a critical operational gap remains: determining whether a newly trained (or adapted) Text2SQL model is production-ready for a specific database before any labels exist. In practice, constructing a labeled test set is often infeasible: data owners may not share labels due to privacy, database schemas may evolve faster than annotation cycles, and manual labeling (drafting SQL, executing, debugging, and verifying) is both costly and time-consuming. Consequently, practitioners are forced into an undesirable decision: deploy without reliable performance estimates or delay deployment to collect labels. Meanwhile, Text2SQL benchmarks are expanding rapidly as new datasets are introduced across increasingly diverse domains, topics, and database structures \cite{li2025omnisql,pham2025multilingual}. This growing diversity exposes Text2SQL models to substantial domain and structural shift, which can cause performance to vary sharply from one test set to another. However, current evaluation practice still lacks a reliable way to measure and quantify such shift in advance, making it difficult to estimate whether a model will generalize well to a new unlabeled test set \cite{jacobsson2025text2sql}. This paper targets that gap directly: \textit{Can we estimate the dataset-level performance of a fixed, well-trained Text2SQL model on an unseen, unlabeled test set?}

As illustrated in \autoref{fig:intro-pipeline}, we introduce a label-free, model-agnostic evaluation framework that predicts performance on new datasets without requiring ground-truth annotations or retraining. To the best of our knowledge, prior works have not tackled label-free, model-agnostic estimation of dataset-level performance for pre-deployment decisions. Existing practices typically rely on annotated test sets or per-example confidence signals rather than dataset-level estimates. Our framework produces a concise reliability report for evaluating Text2SQL models, covering standard Text2SQL metrics, enabling pre-deployment readiness checks, identifying which dataset is most responsible for performance degradation, and reducing dependence on manual labeling. Across diverse domains, schemas, and query complexities, our estimates closely track true performance, offering a scalable path to safer, lower-cost deployment. Our contributions are:

\begin{compactitem}
\item \emph{Problem formulation}: We formalize the label-free, pre-deployment evaluation task for Text2SQL, which is estimating the performance of a fixed, in-service model on unseen, unlabeled data. 

\item \emph{Framework}: We introduce \textbf{FusionSQL}, a model-agnostic evaluator that constructs compact representations of train--test distribution shifts and maps these indicators to dataset-level performance estimates without any model retraining or fine-tuning.

\item \emph{Dataset construction}: We develop \textbf{FusionDataset}, a large-scale and diverse Text2SQL benchmark comprising 3.3M examples, 3.1M unique SQL queries, and 24K databases. It covers extensive schema diversity, multi-dialect SQL structures, and rich linguistic variation with injected distractors, forming the backbone for training and validating FusionSQL.

\item \emph{Empirical validation}: We evaluate across diverse domains, schemas, and query complexities, showing that predicted metrics closely track ground-truth performance, enabling reliable deployment decisions and reducing dependence on costly manual labeling.

\item \emph{Efficiency}: We design the evaluator to be lightweight, using compact shift descriptors and matrix factorization. To support deployment at scale, we optimize shift-descriptor computation, ensuring time efficiency and memory growth.

\end{compactitem}

\section{Related Work}
\label{sec:related_work}

Text2SQL evaluation is traditionally grounded in benchmark datasets with standardized splits (train, test, and validation) and metrics. Spider introduced cross-domain databases with exact-match (EM) and execution accuracy (EX) as the core protocol \cite{yu2018spider}, while SParC and CoSQL extend question types to multi-turn, context-dependent conversations \cite{yu2019sparc,yu2019cosql}. Subsequent benchmarks increase scale and realism, including BIRD \cite{li2024bird}, and Spider~2.0 \cite{lei2025spider}. These datasets define canonical evaluation settings and motivate the transfer scenarios studied in this work. However, all standard protocols assume access to gold SQL and executable databases, which are unavailable in many pre-deployment or client-facing settings. FusionSQL targets this gap by estimating accuracy directly on unlabeled target workloads.


Another line of research studies label-free performance estimation, typically using confidence signals, self-consistency, uncertainty proxies \cite{fu2023estimating,manakul2023selfcheckgpt,farquhar2024detecting}, synthetic evaluation \cite{ren2024autoeval}, or limited labeled subsets via prediction-powered inference \cite{angelopoulos2023ppi}. Classical estimators, such as ATC \cite{garg2022leveraging} and DoC \cite{guillory2021predicting}, originate from other domains (e.g., image classification) and operate based on confidence signals without schema awareness for evaluating Text2SQL models. PseAutoEval \cite{ren2024autoeval} uses pseudo-labeling with human correction to estimate accuracy but requires small sets with labels, limiting its practicality. Judge-based methods, such as BugJudge \cite{liu2025nl2sqlbugs} and ArenaCmp \cite{zheng2023llmasjudge}, use LLMs to assess SQL correctness but incur high computational costs and depend on per-sample judgments. To the best of our knowledge, estimating the accuracy of Text2SQL models on unseen, unlabeled datasets has not been addressed prior to this work. FusionSQL fills this gap by modeling dataset-level train--test shift through pooled embeddings and distributional descriptors, enabling practical accuracy estimation under cross-database shift without labels, executable databases, or per-sample judge calls.

\section{Problem Formulation}
\label{sec:problem_formulation}

\subsection{Model}
We consider a Text2SQL model that maps a natural-language question and a database description to an SQL query. Let $q \in \mathcal{Q}$ be a question, $S$ denote the database schema (including tables, columns, and foreign keys), and $K$ an optional external knowledge base. Following standard formulations:
\begin{equation}
  \hat{y} = f_\psi(q, S, K),
\end{equation} 
where $f_\psi(\cdot)$ is a fixed, trained model with parameters $\psi$, and $\hat{y}$ is the predicted SQL.

At deployment time, the model is applied to a different target environment with questions $q'_i$, schema $S'$, and optional knowledge base $K'$. For a target, unlabeled workload:
\begin{equation}
  \mathcal{D}_T' = \{(q'_i, S', K')\}_{i=1}^{n},
\end{equation}
the model produces predictions:
\begin{equation}
  \hat{y}'_i = f_\psi(q'_i, S', K').
\end{equation}
Let $y_i^\star$ denote the (unknown) gold SQL. With a per-example evaluation function $m(\hat{y}, y^\star, S') \in [0,1]$, the dataset-level performance on $\mathcal{D}_T'$ is:
\begin{equation}
  M^\star = \frac{1}{n}\sum_{i=1}^{n} m(\hat{y}'_i, y_i^\star, S').
\end{equation}
with $m$ being, e.g., exact-match or execution accuracy and $M^\star$ the true (average) accuracy.

\subsection{Challenges}
We target pre-deployment evaluation on unseen, unlabeled data with a shifted environment $(q'_i, S', K')$, which presents the following challenges:
\begin{itemize}
  \item[$\xi1$] \textit{Absence of ground truth}: The labels $\{y_i^\star\}$ are unavailable, so $M^\star$ cannot be computed directly.
  \item[$\xi2$] \textit{Distribution shift}: The target questions, schema, and knowledge $(q'_i, S', K')$ may differ from the training environment $(q, S, K)$ in schema topology, domain vocabulary, linguistic style, and query composition.
  \item[$\xi3$] \textit{Limited access}: The evaluator must operate without modifying the model $f$ or its parameters $\psi$.
  \item[$\xi4$] \textit{Reliability}: Hallucinations or partial correctness can make proxy signals (e.g., train--test distribution shifts) unreliable indicators of true SQL validity.
  \item[$\xi5$] \textit{Efficiency}: Evaluation should be lightweight in wall-clock time, computation, and memory to support routine readiness checks.
\end{itemize}

\subsection{Objective}
The goal is to estimate dataset-level performance on $\mathcal{D}_T'$ without labels and without modifying the Text2SQL model, directly addressing $\xi1$ and $\xi3$. Let $\phi_{\mathrm{src}}(S, K, q)$ denote compact descriptors of the training (or source) environment (schema/knowledge and workload), and let $\phi_{\mathrm{tgt}}(S', K', q')$ denote descriptors of the testing (or target) environment. Define shift descriptors that summarize train-test differences, which addresses $\xi2$:
\begin{equation}
  \Delta \;=\; h \big(\phi_{\mathrm{tgt}}(S', K', q'), \phi_{\mathrm{src}}(S, K, q)\big).
\end{equation}
We seek an evaluator $g$, parameterized by $\theta$ and driven solely by shift descriptors under label scarcity:
\begin{equation}
  \widehat{M} \;=\; g_\theta(\Delta),
\end{equation}
and satisfies the following properties, each addressing the challenges outlined above:
\begin{itemize}
  \item \textit{Accuracy} ($\xi1$): minimize $|\widehat{M} - M^\star|$ and mean-squared error across target workloads.
  \item \textit{Uncertainty} ($\xi2$, $\xi4$): provide calibrated uncertainty estimates, e.g., a prediction interval $[\widehat{M}-\delta_\alpha,\; \widehat{M}+\delta_\alpha]$ such that $\mathbb{P}(M^\star \in [\widehat{M}-\delta_\alpha,\; \widehat{M}+\delta_\alpha]) \ge 1-\alpha,$ where $\delta_\alpha$ is the interval half-width at miscoverage level $\alpha$.
  \item \textit{Efficiency} ($\xi5$): operate with low runtime and resource usage to enable rapid pre-deployment decisions.
  \item \textit{Generality} ($\xi3$): require no access to ground-truth SQL, no populated databases for execution, and no changes to $f$ or $\psi$.
\end{itemize}

\section{FusionSQL Framework}
\label{sec:methodology}
\begin{figure*}[t]
  \centering
  \includegraphics[width=\linewidth]{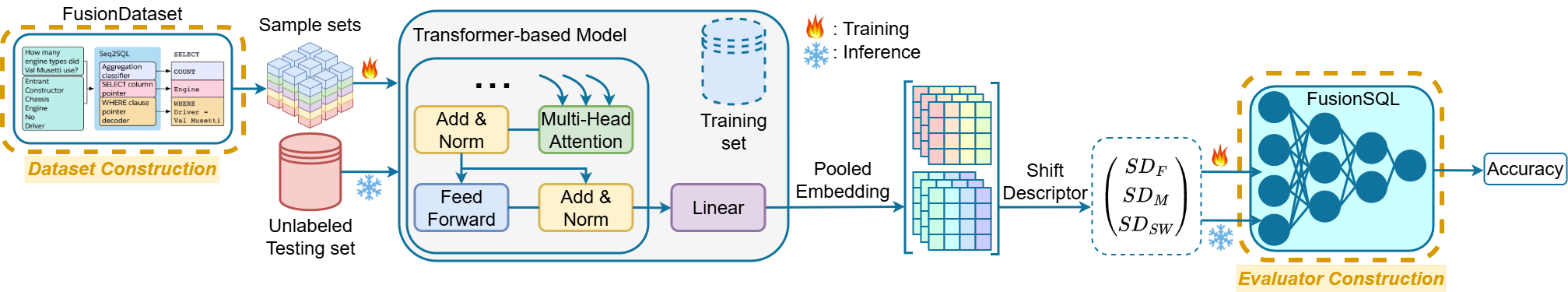}
  \vspace{-1.5em}
  \caption{{\textbf{FusionSQL framework.} \textit{Training}: A frozen Text2SQL model encodes training and FusionDataset samples into embeddings to compute shift descriptors (\(\textit{SD}_F\), \(\textit{SD}_M\), \(\textit{SD}_{SW}\)) for training the FusionSQL evaluator. \textit{Inference}: For unseen, unlabeled workloads, the same descriptors are computed to estimate accuracy without labels or retraining.}}
  \label{fig:fusionsql_framework}
  \vspace{-1.5em}
\end{figure*}

To address the challenges and objectives outlined in \autoref{sec:problem_formulation}, we propose \textbf{FusionSQL}, a label-free, model-agnostic evaluator that estimates dataset-level performance on unseen workloads using only train--test shift descriptors, without target labels or model retraining. As shown in \autoref{fig:fusionsql_framework}, the proposed approach is presented in two logical components for clarity: \textit{Dataset Construction}, which builds representative shift scenarios, and \textit{Evaluator Construction}, which learns the function $g_\theta(\Delta)$ for performance prediction.

\subsection{Dataset Construction}
\label{sec:dataset_construction}

To train a robust and generalizable evaluator (\autoref{sec:evaluator_construction}), we construct a large-scale, realistic Text2SQL corpus, FusionDataset, designed to cover three aspects: (1) database diversity, (2) SQL structural diversity, and (3) natural-language question diversity with distractors. By jointly covering these aspects, FusionDataset exposes FusionSQL to a wide spectrum of schema, query, and linguistic shifts, enabling stable accuracy estimation under unseen train--test conditions.

\sstitle{Database Diversity}
We collect tables from TabLib \cite{eggert2023tablib} and KaggleDBQA \cite{lee-2021-kaggle-dbqa}, spanning business, education, healthcare, and government domains. After language, size, duplication, and semantic filtering (using GPT-5), tables are clustered by schema and topic, and normalized to form relational databases.

\sstitle{SQL Structural Diversity}
From these databases, we synthesize queries across simple to highly complex regimes using OmniSQL \cite{li2025omnisql}, augmented with SQLForge \cite{guo2025sqlforge}, PARSQL \cite{dai2025parsql}, and operator-rewriting strategies \cite{cui2025llm}. This process yields 9.4M SQL queries (8.1M unique after canonicalization) spanning 42 operators, 8 nesting depths, and 12 dialect variants, substantially exceeding existing benchmarks in scale and variety. The generated SQLs are subsequently filtered using execution-based checks and feedback from independent judges, such as GPT-5, to retain only high-quality candidates.

\sstitle{Question Diversity and Distractors}
Each SQL is paired with multiple natural-language questions covering diverse styles \cite{li2025omnisql}. We inject distractors and irrelevant modifiers, as observed in KaggleDBQA and BIRD \cite{li2024bird}, to model noisy user inputs and schema-irrelevant content. For example, in a schema with tables \texttt{customer} and \texttt{orders}, a user might ask: \textit{``List the top 10 coolest customers who recently went viral on TikTok, with emails.''}. The gold SQL is: \texttt{SELECT c.name, c.email FROM customer AS c WHERE c.email IS NOT NULL ORDER BY c.customer\_id LIMIT 10;}. Here, \textit{``went viral on TikTok''} is a distractor--natural in phrasing but semantically irrelevant--while \textit{``with emails''} dictates the actual constraint.

Overall, FusionDataset significantly expands existing Text2SQL corpora in database count, SQL variety, and linguistic coverage, forming the backbone of FusionSQL and enabling principled evaluation under realistic schema, query, and language shifts. Further experiments validating its coverage are presented in \autoref{sec:data_coverage_experiment}.
\subsection{Evaluator Construction}
\label{sec:evaluator_construction}

Given the objective in \autoref{sec:problem_formulation}, FusionSQL aims to approximate the true performance $M^\star$ through a predictive function $g_\theta(\cdot)$ conditioned on the train--test shift descriptors $\Delta$.
Formally, we model:
\begin{equation}
\widehat{M} = g_\theta(\Delta),
\label{eq:evaluator_model}
\end{equation}
where $\theta$ are learnable parameters of the evaluator.
The goal is to minimize the mean squared deviation between the predicted and true dataset--level metrics:
\begin{equation}
\min_{\theta} \; \mathbb{E}\big[(\widehat{M} - M^\star)^2\big]
   = \mathbb{E}\big[(g_\theta(\Delta) - M^\star)^2\big].
\label{eq:evaluator_objective}
\end{equation}

We inherit the pipeline of CHESS-SQL \cite{talaei2024chess} for fine-tuning a Text2SQL model $f_\psi(\cdot)$ until it achieves the best validation accuracy. Later, we train $g_\theta(\Delta)$ on a labeled meta-collection of workloads so that it learns to map observed train--test differences (captured by $\Delta$) to their corresponding accuracies. FusionSQL is implemented as a 3-layer MLP, with architectural and optimization details provided in \autoref{sec:experiments} and its selection over alternative regressors justified through ablations in \autoref{sec:ablation}.

\sstitle{Training phase}
FusionSQL learns to map distributional differences between training and deployment environments $\Delta$ to downstream model performance. Let $\phi_{\mathrm{train}}$ denote the original training environment of the Text2SQL model, and let $\phi_{\mathrm{syn}}$ denote a synthetic target environment drawn from the FusionDataset in \autoref{sec:dataset_construction}. We summarize the train--test difference between these two environments using a shift descriptor:
\begin{equation}
\Delta_{\text{train}} = h(\phi_{\mathrm{train}}, \phi_{\mathrm{syn}}),
\label{eq:delta_train}
\end{equation}
where $h(\cdot)$ denotes a deterministic function that extracts a fixed-length representation of distributional shift between environments. FusionSQL, or $g_\theta(\cdot)$, is trained so that, given $\Delta_{\text{train}}$, it predicts the observed dataset-level execution accuracy of the Text2SQL model on the synthetic target environment. Formally, training minimizes the discrepancy between $g_\theta(\Delta_{\text{train}})$ and the true execution accuracy $M^\star_{\text{syn}}$ of a Text2SQL model on the FusionDataset, as in \autoref{eq:evaluator_objective}. The construction of $h$ is independent of model parameters and does not require labeled target data.

\sstitle{Shift Descriptors} To capture this train--test difference (\autoref{eq:delta_train}), FusionSQL utilizes a shift descriptor that summarizes complementary aspects of distributional change between two Text2SQL datasets. Given pooled embedding sets from the model's training workload $\mathcal{S}=\{\phi_i\}_{i=1}^{n}$ and from a target workload $\mathcal{T}=\{\phi'_j\}_{j=1}^{m}$, where $\phi_i,\phi'_j\in\mathbb{R}^D$, where $D$ denotes the embedding dimensionality of each pooled representation vector. We instantiate $\Delta_{\text{train}} = h(\phi_{\mathcal{S}},\phi_{\mathcal{T}}) = [\textit{SD}_F(\mathcal{S},\mathcal{T}),\textit{SD}_M(\mathcal{S},\mathcal{T}),\textit{SD}_{SW}(\mathcal{S},\mathcal{T})]$. At deployment, the same construction is applied to the real target workload, $\Delta_{\text{test}} = h(\phi_{\mathcal{S}},\phi_{\mathrm{test}})$, using pooled embeddings from the original training data and the unseen test set. Here $\textit{SD}_F$, $\textit{SD}_M$, and $\textit{SD}_{SW}$ capture complementary changes in Text2SQL workloads, reflecting shifts in (i) schema complexity and query composition through global embedding statistics, (ii) rare or failure-prone cases via tail behavior, and (iii) structural reorganization of question--schema interactions through distributional shape, since last-layer pooled embeddings encode linguistic inputs with their alignment to database schemas.

\begin{compactitem}
    \item The Fr\'echet descriptor $\textit{SD}_F$ captures \textit{global} domain drift by comparing first- and second-order statistics of the two embedding sets: $\mu_{\mathcal{S}}=\frac{1}{n}\sum_{i=1}^{n}\phi_i$ and $\mu_{\mathcal{T}}=\frac{1}{m}\sum_{j=1}^{m}\phi'_j$ denote the element-wise means, while $\sigma^2_{\mathcal{S}},\sigma^2_{\mathcal{T}}\in\mathbb{R}^D$ denote the corresponding element-wise variances. In particular, $\textit{SD}_F$ includes the mean-shift term $\|\mu_{\mathcal{T}}-\mu_{\mathcal{S}}\|_2^2$ and variance-ratio terms, and becomes large under systematic changes such as moving from single-table factual queries to multi-join or nested SQL. 
    \item The Mahalanobis descriptor $\textit{SD}_M$ emphasizes \textit{tail behavior} by whitening each target embedding with the source statistics, $r_j=\|(\phi'_j-\mu_{\mathcal{S}})\odot\sigma_{\mathcal{S}}^{-1}\|_2$, and summarizing deviations via $\textit{SD}_M=[\mu_r,\sigma_r]$. This highlights rare or atypical queries (e.g., unusual aggregations or conversational follow-ups) that cause failure cases under shift. 
    \item The Sliced Wasserstein Distance $\textit{SD}_{SW}$ detects \textit{distributional shape} changes by projecting embeddings onto $L$ directions and averaging one-dimensional $W_2$ distances between sorted projections \cite{viethuynh2021,nhatho2019pmlr,nhatho2017icml}. It is particularly sensitive to directional distortions caused by schema restructuring or collapsing query templates across datasets. 
\end{compactitem} 
Together, $\textit{SD}_F$, $\textit{SD}_M$, and $\textit{SD}_{SW}$ provide complementary signals: global drift, tail risk, and structural reorganization. Since $\textit{SD}_{SW}$ is most computationally expensive, we discuss a possible optimization in \autoref{sec:further_consideration}. In addition, \autoref{sec:ablation} presents illustrative Text2SQL transfers and examples that demonstrate the value of this combination, and contrasts it with the commonly used Euclidean distance, which shows substantially weaker correlation with accuracy under distribution shift.

\sstitle{Testing phase}
At inference time, ground-truth labels are unavailable.
We only need to compute the real train--test difference $\Delta_{\text{test}}$. Passing $\Delta_{\text{test}}$ into $g_\theta(\cdot)$ yields:
\begin{equation}
\widehat{M}_{\text{test}} = g_\theta(\Delta_{\text{test}}),
\end{equation}
which corresponds to the predicted execution accuracy of the Text2SQL model on new data.
This process requires neither ground-truth SQL queries nor retraining of the base model, enabling a label-free evaluation in real-world deployments.

\subsection{Further Considerations}
\label{sec:further_consideration}

\sstitle{Hybrid SWD}
\label{sec:swd_design}
FusionSQL uses sliced Wasserstein distance (SWD) to capture directional and non-Gaussian shifts between source and target embedding sets. Its runtime grows linearly with the number of projection directions $L$:
\begin{equation}
O\big(L[D(n+m)] + L[(n+m)\log(n+m)]\big)
\label{eq:swd_complexity}
\end{equation}
where $n,m$ are the numbers of source and target embeddings from the Text2SQL model, and $D$ is the embedding dimension. The first term is projection cost, and the second term is sorting cost. To reduce this overhead, we adopt a \textbf{Hybrid SWD} scheme that combines data-aware and random projections: (i) we compute the top-$k$ principal components (typically $k{=}8$--$10$) from a subsample of the joint embedding space $\mathcal{S}\cup\mathcal{T}$ using randomized PCA at cost $O((n{+}m)Dk)$; (ii) we append $R{=}16$--$24$ random unit directions to obtain $L{=}k{+}R\approx24$--$34$ slices, compared to $L\approx64$ in a fully random SWD; and (iii) we vectorize all projections as a single matrix multiplication and reuse cached projections across batches to reduce memory movement. The hybrid configuration preserves fidelity while substantially reducing latency and memory. The overall complexity:
\begin{equation}
O\big((n+m)Dk + L[(n+m)\log(n+m)]\big),
\label{eq:hybrid_swd_complexity}
\end{equation}
which provides a practical upper bound on runtime and memory usage during evaluation. Experimental results in \autoref{sec:ablation} confirm that Hybrid SWD maintains comparable accuracy to all-random SWD while reducing latency and memory.

\sstitle{Generalization to Unseen Models} To enable FusionSQL to generalize to \textit{unseen} Text2SQL models, we adopt a meta-learning strategy \cite{nichol2018reptile} that learns a model-agnostic evaluator initialization optimized for rapid adaptation to novel architectures. Meta-training spans a diverse pool of Text2SQL systems with shared backbone families and varied designs, encouraging aligned internal representations and shift patterns. We treat each base Text2SQL model $m$ in the meta-training pool as a meta-task, where multiple sampled subsets $S_i\subseteq\textit{FusionDataset}$ (\autoref{sec:data_efficiency_experiment}) induce distinct shift conditions within the same task. FusionDataset is constructed as a synthetic meta-set spanning diverse schema structures, SQL operators, and linguistic patterns, allowing the same $S_i$ to be reused at deployment without requiring labeled or partially labeled target data. Given many sampled subsets $S_i$, FusionSQL covers a wide spectrum of $(m,S_i)$ shift conditions and yields an initialization that adapts effectively after only a few gradient steps, as formalized in \cite{finn2017model}. For each $(m,S_i)$ pair, we compute pooled embeddings $\phi_{\text{train}}$ for the model's training workload and $\phi_{S_i}$ for the sampled subset, and derive shift descriptors $\Delta_{m,S_i}=h\!\left(\phi_{\text{train}},\phi_{S_i}\right)$, where $h(\cdot)$ denotes FusionSQL's descriptor mapping. An evaluator $g_\theta$ predicts the accuracy of model $m$ on $S_i$, yielding the task-level loss:
\begin{equation}
\mathcal{L}_{m,S_i}(\theta)
=
\ell\!\left(g_\theta(\Delta_{m,S_i}),\mathrm{Acc}_{m,S_i}\right),
\label{eq:meta_loss}
\end{equation}
with $\ell$ the error objective and $\mathrm{Acc}_{m,S_i}$ the corresponding accuracy of model $m$ on subset $S_i$. Following Reptile \cite{nichol2018reptile}, we learn an evaluator initialization $\theta$ that performs well after a small number of adaptation steps on a new model, rather than minimizing average error over the meta-training pool. For each meta-task $m$ we apply a few inner updates: $\theta_m = \theta - \alpha \nabla_\theta \mathbb{E}_{S_i}\mathcal{L}_{m,S_i}(\theta)$, and update the initialization toward the adapted parameters via: $\theta \leftarrow \theta + \epsilon(\theta_m-\theta)$, where $\alpha$ and $\epsilon$ are the inner- and outer-loop step sizes. At deployment, for a previously unseen model $m_{\text{new}}$, we adapt the evaluator using $S_i$. Since $S_i$ is known, we can compute the model's accuracy and form the loss $\mathcal{L}_{m_{\text{new}},S_i}(\theta)$. We then perform gradient-based adaptation steps \cite{finn2017model} to obtain model-specific parameters $\theta_{m_{\text{new}}}$. Target evaluation still remains fully label-free and we predict accuracy on an unseen workload using $\Delta_{m_{\text{new}},\text{test}} = h\!\left(\phi_{\text{train}},\phi_{\text{test}}\right)$, where $\phi_{\text{train}}$ and $\phi_{\text{test}}$ are the pooled embedding distributions of $m_\text{new}$'s training and unlabeled target workloads, respectively. This enables a fast and model-specific evaluator $g_{\theta_{m_{\text{new}}}}$ without requiring labels on the target workload (\autoref{sec:ablation}).

\section{Experiments}
\label{sec:experiments}

Designing a label-free evaluator that predicts Text2SQL model performance from distributional evidence raises several research questions (RQs) for empirical evaluation. 
\begin{compactitem}
    \item[\textbf{RQ1 (Data Coverage)}\hfill(\autoref{sec:data_coverage_experiment}):]
    How well does FusionDataset capture the diversity of real-world Text2SQL workloads with respect to schema structures, SQL operator usage, and natural--language patterns?

    \item[\textbf{RQ2 (Data Efficiency)} \autoref{sec:data_efficiency_experiment}:]
    How do the size and composition of FusionDataset influence FusionSQL's ability to generalize across datasets under data constraints?

    \item[\textbf{RQ3 (Evaluator Learning)} \autoref{sec:evaluator_learning_experiment}:]
    How well does FusionSQL estimate accuracy from shift descriptors and generalize across unseen datasets and Text2SQL models?

    \item[\textbf{RQ4 (Scalable Deployment)} \autoref{sec:ablation}:]
    How can we optimize FusionSQL to remain accurate while scaling to large workloads and frequent evaluation cycles?

    \item[\textbf{RQ5 (Shift Modeling)} \autoref{sec:ablation}:]
    How can we summarize train--test discrepancies between two Text2SQL workloads for both neural and classical systems?
\end{compactitem}

\begin{table*}[t]
\centering
\footnotesize
\caption{\textbf{Comparison of Text2SQL datasets.}
\textit{\# Examples} count question--SQL pairs, while \textit{\# Unique SQL} counts distinct canonicalized SQL programs. $^{\star}$ Each WikiSQL ``DB'' contains a single table; $\dagger$ BIRD's Unique SQL is unavailable due to its inaccessible test set. FusionDataset offers greater structural and linguistic diversity than prior datasets.}
\label{tab:overall_stats}
\vspace{-0.5em}
\resizebox{0.65\textwidth}{!}{%
\begin{tabular}{l l r r r}
\toprule
\textbf{Dataset} & \textbf{Source} & \textbf{\# Examples} & \textbf{\# Unique SQL} & \textbf{\# DB}\\
\midrule
WikiSQL \cite{zhong2017seq2sql} & Human+Template & 80{,}654 & 80{,}257 & 26{,}531$^{\star}$ \\
Spider \cite{yu2018spider} & Human & 10{,}181 & 5{,}693 & 200 \\
SParC \cite{yu2019sparc} & Human & 12{,}726 & --- & 200 \\
CoSQL \cite{yu2019cosql} & Human & 10{,}000{+} & --- & 200 \\
BIRD \cite{li2024bird} & Human & 12{,}751 & ---$\dagger$ & 95 \\
ScienceBenchmark \cite{stockinger2023sciencebenchmark} & LLM-Gen+Human+Template & 5{,}031 & 3{,}652 & 3 \\
EHRSQL \cite{lee2022ehrsql} & Human+Template & 20{,}108 & 18{,}253 & 2 \\
KaggleDBQA \cite{lee-2021-kaggle-dbqa} & Human & 272 & --- & 8 \\
SynSQL-2.5M \cite{li2025omnisql} & LLM-Gen & 2{,}544{,}390 & 2{,}412{,}915 & 16{,}583 \\
\textbf{FusionDataset (ours)} & LLM-Gen & 3{,}373{,}204 & 3{,}119{,}472 & 24{,}625 \\
\bottomrule
\end{tabular}
}
\vspace{-1em}
\end{table*}

\sstitle{Evaluation Datasets}
At test time, FusionSQL is evaluated on seven established Text2SQL benchmarks spanning diverse domains, query complexities, and interaction settings. These include Spider \cite{yu2018spider}: 10.1K cross-domain questions over 200 databases; WikiSQL \cite{zhong2017seq2sql}: 80K questions over single-table databases; BIRD \cite{li2024bird}: 12.7K realistic business and research queries requiring numeric grounding; SParC \cite{yu2019sparc} and CoSQL \cite{yu2019cosql}: multi-turn conversational parsing and clarification; SynSQL-2.5M \cite{li2025omnisql}: a large-scale synthetic dataset covering 16K databases and multiple SQL dialects; and Spider~2.0 \cite{lei2025spider}: an enterprise-scale extension of Spider with real-world databases and substantially deeper query nesting.

\sstitle{Evaluation Metrics}
We report standard Text2SQL accuracy metrics, including Exact Match (EM) and Execution Accuracy (EX), which are used in \cite{yu2018spider,li2024bird,jacobsson2025text2sql}:
\begin{equation}
\mathrm{EM}=\frac{1}{N}\sum_{i=1}^{N}\mathbb{I}[\hat{y}_i=y_i],\qquad
\mathrm{EX}=\frac{1}{N}\sum_{i=1}^{N}\mathbb{I}[\hat{y}_i\equiv y_i]_S,
\label{eq:em_ex}
\end{equation}
where $N$ is the number of examples, where $\hat{y}_i$ and $y_i$ denote the predicted and gold SQL for the example $i$, respectively; $S$ is the corresponding database schema; $\equiv$ denotes equivalence under execution. The effectiveness of FusionSQL is quantified using MAE~\cite{zheng2023gnnevaluator} between predicted and actual accuracy:
\begin{equation}
\mathrm{MAE} = \frac{1}{N} \sum_{i=1}^{N} | \hat{y}_i - y_i |,
\label{eq:mae}
\end{equation}
where $\hat{y}_i$ denotes the predicted accuracy for dataset $i$, and $y_i$ is the corresponding true accuracy. A lower MAE indicates a more reliable and stable evaluator. We report EM/EX for model accuracy and MAE for evaluator accuracy across all transfers to measure generalization.

\sstitle{Base Models}
Shift descriptors are collected from five representative Text2SQL systems: Qwen2.5-72B-Instruct \cite{yang2025qwen3}, Llama-3.1-70B-Instruct \cite{grattafiori2024llama}, DeepSeek-Coder-33B \cite{guo2024deepseek}, XiYanSQL-14B\footnote{XiYanSQL-14B stands for XiYanSQL-QwenCoder-14B.} \cite{XiYanSQL}, and CSC-SQL-7B\footnote{CSC-SQL-7B stands for CscSQL-Grpo-Qwen2.5-Coder-7B-Instruct.} \cite{sheng2025csc}.  
Each model is trained on its original training set and selected based on its validation performance. After training, the model is frozen. We then pair its original training set with multiple subsets sampled from FusionDataset, and extract the model's latent representations from each pair to construct shift descriptors.

\sstitle{Training Settings}
FusionSQL is implemented as a three-layer MLP with hidden dimensions \{256, 128, 64\}, ReLU activations, and layer normalization, and is trained to regress the true accuracy of a Text2SQL model from shift descriptors. Training uses a batch size of 64 and the AdamW optimizer with learning rate $1{\times}10^{-4}$ (cosine decay), $\beta_1{=}0.9$, $\beta_2{=}0.999$, and weight decay $1{\times}10^{-3}$. The model is trained for up to 20 epochs with early stopping based on validation MAE, applies dropout of 0.2 between hidden layers, and optimizes mean squared error (MSE) between predicted and true accuracies.

\sstitle{Environment}
All experiments are conducted on a workstation equipped with two NVIDIA GeForce RTX~4090 GPUs (24\,GB each) and an Intel(R) Core(TM) i7-14700 CPU (20 cores, 2.1\,GHz base clock). Computation is performed with mixed-precision (\texttt{bfloat16}) and parallel data loading for efficient descriptor extraction and evaluator training.

\subsection{Data Coverage}
\label{sec:data_coverage_experiment}

To answer \textbf{RQ1 (Data Coverage)}, we assess whether FusionDataset provides comprehensive coverage of real Text2SQL workloads in terms of schema design, SQL composition, and natural--language usage.

\sstitle{Data Size}
\autoref{tab:overall_stats} compares FusionDataset with nine public Text2SQL benchmarks: WikiSQL \cite{zhong2017seq2sql}, Spider \cite{yu2018spider}, SParC \cite{yu2019sparc}, CoSQL \cite{yu2019cosql}, BIRD \cite{li2024bird}, ScienceBenchmark \cite{stockinger2023sciencebenchmark}, EHRSQL \cite{lee2022ehrsql}, KaggleDBQA \cite{lee-2021-kaggle-dbqa}, and SynSQL-2.5M \cite{li2025omnisql}. FusionDataset contains 3{,}373{,}204 question-SQL pairs across 24{,}625 databases, substantially larger than most benchmarks in both examples and database count. It features 3{,}119{,}472 unique SQL queries after canonicalization, indicating low redundancy and broad structural coverage. This scale and variety provide a rich foundation for modeling distributional shifts.

\sstitle{Difficulty Analysis}
We further verify that FusionDataset provides comprehensive coverage across difficulty levels. Following \cite{yu2018spider}, examples are categorized into four tiers--\textit{Simple} (30\%), \textit{Moderate} (35\%), \textit{Complex} (25\%), and \textit{Highly Complex} (10\%)--based on SQL length, join count, and operator composition. \autoref{tab:difficulty_levels} reports per-tier accuracy metrics. The monotonic performance decline across tiers demonstrates that FusionDataset accurately captures a realistic gradient of reasoning complexity.

\begin{table*}[h]
\centering
\footnotesize
\caption{\textbf{Difficulty-stratified results.} EM and EX (\%) decrease consistently with increasing complexity, confirming coverage of a realistic reasoning spectrum.
}
\vspace{-0.5em}
\setlength{\tabcolsep}{5.5pt}
\resizebox{0.7\textwidth}{!}{%
\begin{tabular}{lcccccccc|cc}
\toprule
\multirow{2}{*}{Model} &
\multicolumn{2}{c}{Simple (30\%)} &
\multicolumn{2}{c}{Moderate (35\%)} &
\multicolumn{2}{c}{Complex (25\%)} &
\multicolumn{2}{c|}{Highly Complex (10\%)} &
\multicolumn{2}{c}{Overall} \\
\cmidrule(lr){2-3}\cmidrule(lr){4-5}\cmidrule(lr){6-7}\cmidrule(lr){8-9}\cmidrule(lr){10-11}
 & EM & EX & EM & EX & EM & EX & EM & EX & EM & EX \\
\midrule
Qwen2.5-72B-Instruct   & 82.1 & 86.5 & 74.5 & 80.1 & 63.2 & 70.5 & 51.4 & 59.8 & 71.6 & 77.2 \\
Llama-3.1-70B-Instruct & 80.4 & 84.9 & 72.3 & 78.6 & 61.1 & 68.3 & 49.5 & 56.7 & 69.7 & 74.6 \\
DeepSeek-Coder-33B     & 75.2 & 80.8 & 67.8 & 73.9 & 56.0 & 62.5 & 44.2 & 52.0 & 64.7 & 70.3 \\
CSC-SQL-7B             & 68.8 & 74.1 & 60.1 & 66.5 & 48.7 & 55.2 & 37.3 & 44.0 & 57.6 & 63.0 \\
XiYanSQL-14B           & 71.0 & 76.4 & 63.2 & 69.8 & 52.0 & 58.6 & 40.5 & 47.2 & 60.5 & 66.1 \\
\bottomrule
\end{tabular}
}
\label{tab:difficulty_levels}
\vspace{-1em}
\end{table*}

\begin{figure}[t]
  \centering
  \includegraphics[width=0.9\linewidth]{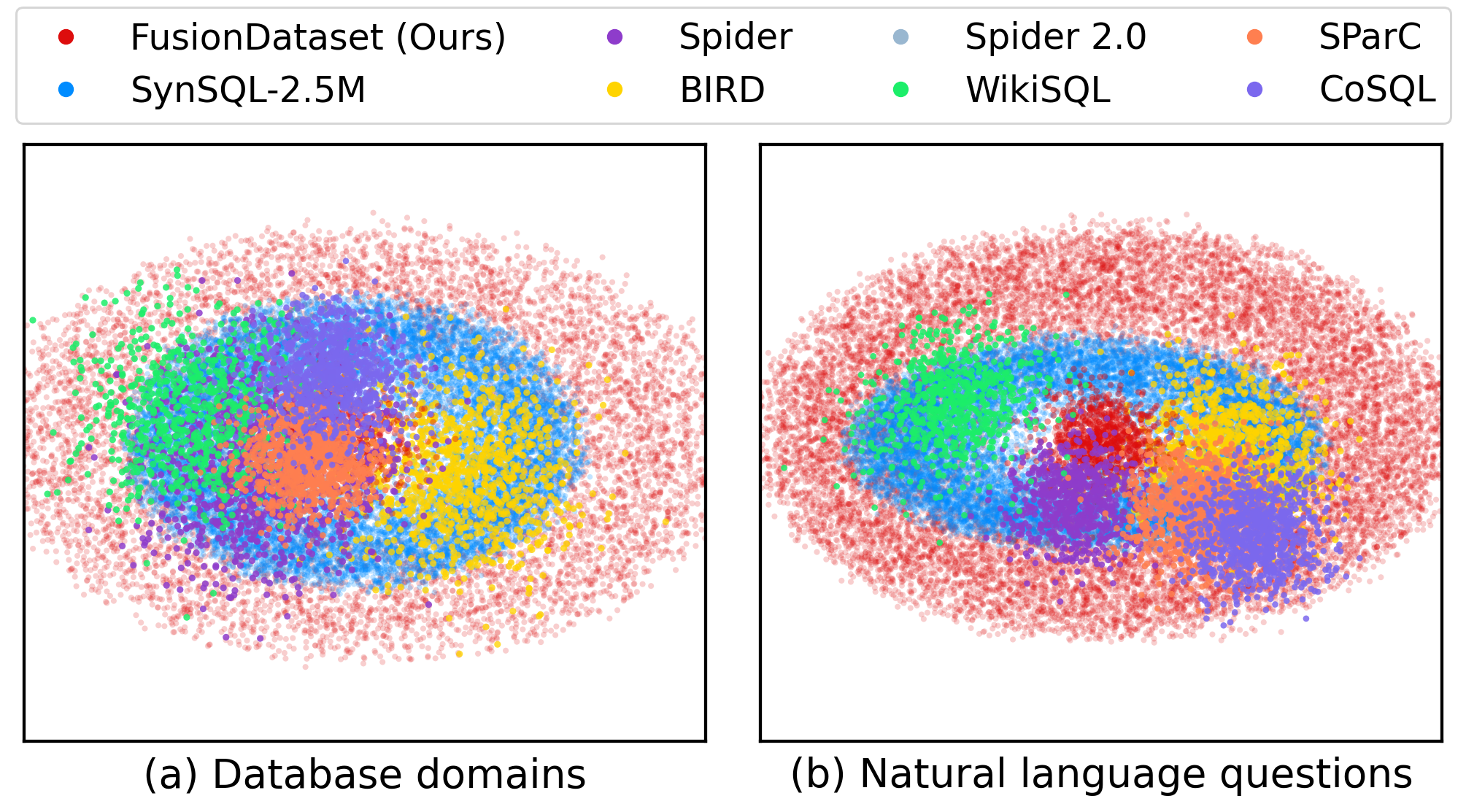}
  \caption{\textbf{t-SNE coverage.} Comparing 50K samples, FusionDataset bridges clusters of existing benchmarks in both domain (a) and question space (b), reflecting broader semantic and structural diversity of real-world Text2SQL variability.}
  \label{fig:tsne-coverage}
  \vspace{-.5em}
\end{figure}

\sstitle{Semantic Coverage}
\autoref{fig:tsne-coverage} visualizes two t-SNE projections comparing 50K samples from FusionDataset and seven public Text2SQL benchmarks: SynSQL-2.5M \cite{li2025omnisql}, Spider \cite{yu2018spider}, BIRD \cite{li2024bird}, Spider~2.0 \cite{lei2025spider}, WikiSQL \cite{zhong2017seq2sql}, SParC \cite{yu2019sparc}, and CoSQL \cite{yu2019cosql}. FusionDataset (red) exhibits the broadest footprint, overlapping all regions and forming a continuous bridge between synthetic and real-world data. SynSQL-2.5M (blue) covers a wide but uniform area due to synthetic schema diversity, while Spider and Spider~2.0 (purple, gray) occupy dense clusters from realistic multi-table databases. WikiSQL (green) appears compact because of its single-table structure, and BIRD (yellow) extends toward value-grounded domains. SParC and CoSQL (orange, violet) cluster near Spider but shift along a conversational axis. Overall, FusionDataset unifies these regions, showing the widest semantic and linguistic coverage across domains and question styles.

\begin{figure}[t]
  \centering
  \scalebox{1}[0.9]{\includegraphics[width=0.8\linewidth]{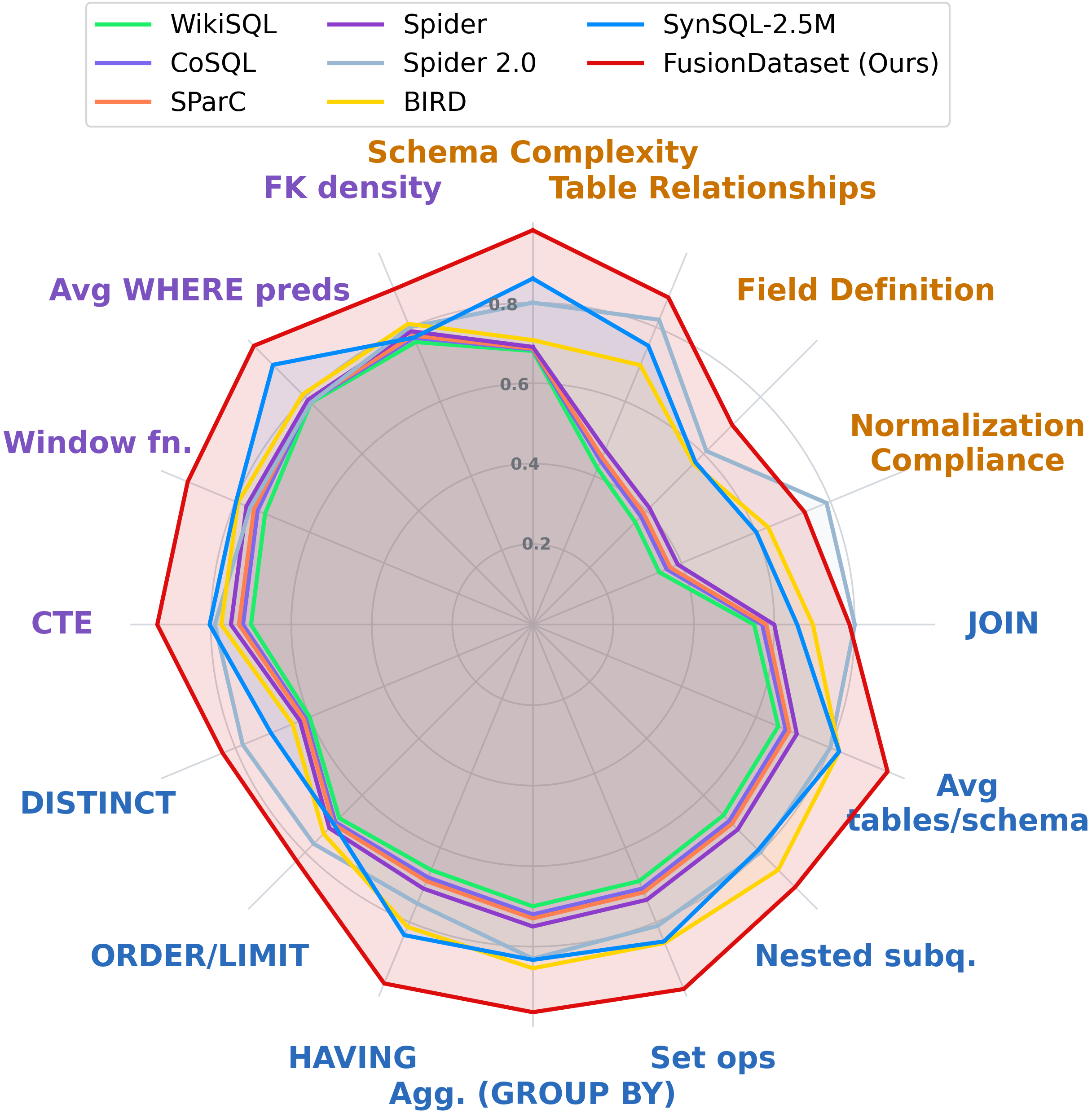}}
  \caption{\textbf{Structural coverage.} Radar chart over 16 schemas and SQL complexity dimensions shows that FusionDataset (red) consistently achieves  higher normalized coverage than existing datasets, reflecting diverse schema and query structures.}
  \label{fig:radar_coverage}
  \vspace{-1em}
\end{figure}

\sstitle{Structural Coverage}
\autoref{fig:radar_coverage} compares seven Text2SQL datasets across 16 normalized schema and query complexity dimensions, capturing nesting, set operations, aggregation, schema connectivity, and predicate density. FusionDataset achieves the broadest coverage, reflecting the richest schemas and most diverse queries, while SynSQL-2.5M also spans a wide range due to synthetic generation. Spider and Spider~2.0 provide strong mid-level baselines; BIRD, SParC, and CoSQL show lower schema complexity with moderate operator diversity, and WikiSQL remains the simplest benchmark. Overall, FusionDataset extends the structural envelope of existing datasets, supporting cross-dataset generalization.

\subsection{Data Efficiency}
\label{sec:data_efficiency_experiment}

This section investigates \textbf{RQ2 (Data Efficiency)} by studying how FusionSQL performs under limited budgets and constraints. In practice, constructing large--scale training corpora is often infeasible due to financial and computational constraints. We therefore examine how different strategies for forming sample sets and meta-sets from FusionDataset affect cross--dataset generalization.

\begin{figure}[!h]
    \centering
    \includegraphics[width=0.8\linewidth]{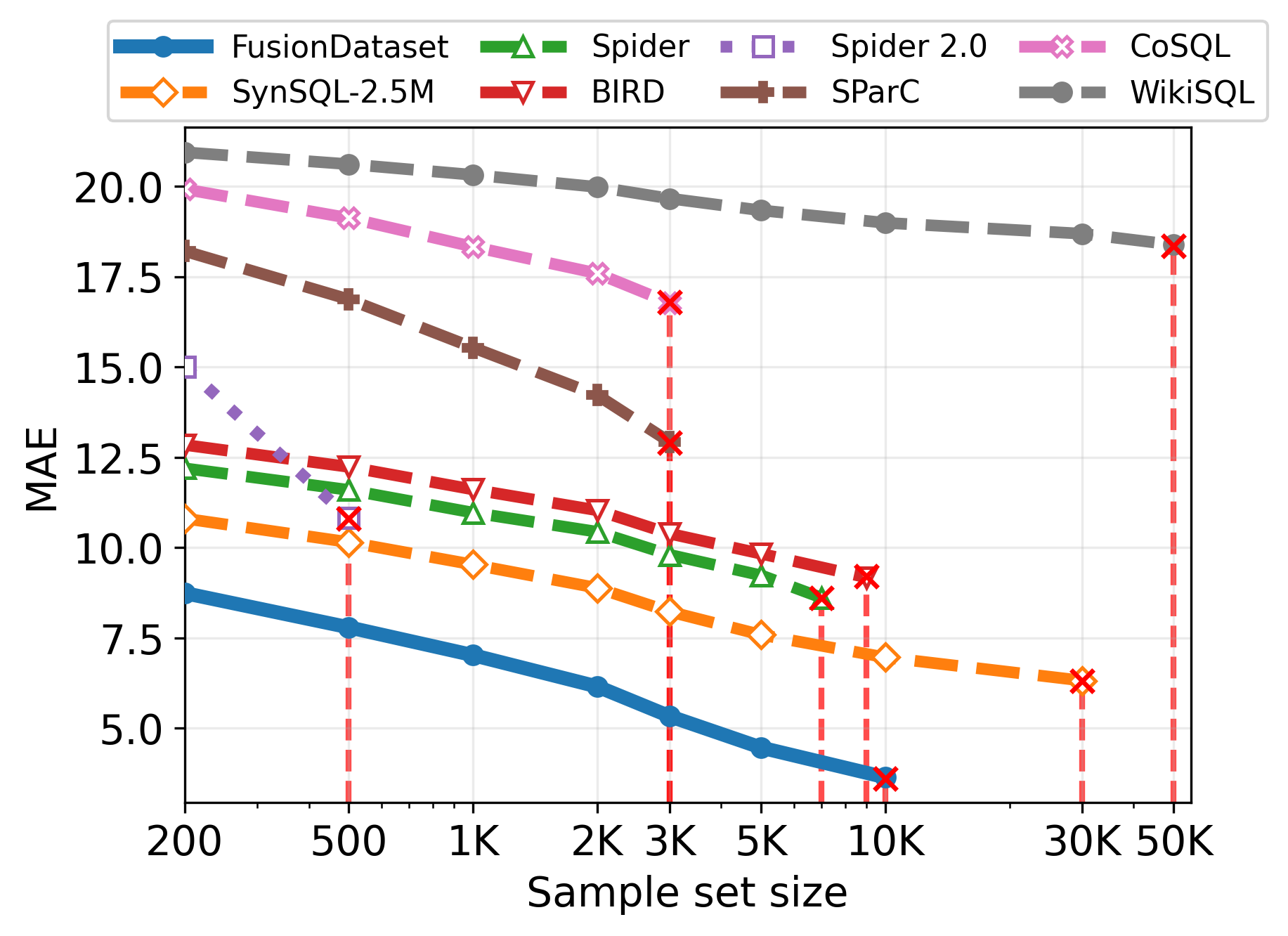}
    \caption{\textbf{Sample-set size.} Impact of sample-set size $|S_i|$ used to compute distribution shifts for an instance $(\mathcal{D}_{\mathrm{train}},S_i)$. Error decreases as $|S_i|$ grows.}
    \label{fig:fusiondataset_samplesize}
    \vspace{-2em}
\end{figure}

\begin{table*}[h]
\centering
\footnotesize
\setlength{\tabcolsep}{3.6pt}
\caption{MAE ($\downarrow$) of dataset-level accuracy estimation for various source-target transfers. Each cell reports mean $\pm$ 95\% CI (percentage points). Best in \textbf{bold}, second-best \underline{underlined}.}
\label{tab:mae-evaluator}
\vspace{-.5em}
\resizebox{0.8\textwidth}{!}{%
\begin{tabular}{llccccc c}
\toprule
\textbf{Transfer ($A \rightarrow B$)} & \textbf{Methods} &
\textit{Qwen2.5-72B} &
\textit{Llama-3.1-70B} &
\textit{DeepSeek-33B} &
\textit{XiYanSQL-14B} &
\textit{CSC-SQL-7B} &
\textit{Avg.} \\
\midrule
\multirow{10}{*}{Spider $\rightarrow$ BIRD}
& ATC-MC \cite{garg2022leveraging} & 13.9 $\pm$ 1.1 & 14.6 $\pm$ 1.2 & 15.2 $\pm$ 1.2 & 17.4 $\pm$ 1.4 & 18.3 $\pm$ 1.5 & 15.9 $\pm$ 1.3 \\
& ATC-NE \cite{garg2022leveraging} & 15.0 $\pm$ 1.2 & 15.7 $\pm$ 1.3 & 16.5 $\pm$ 1.3 & 18.6 $\pm$ 1.5 & 19.8 $\pm$ 1.6 & 17.1 $\pm$ 1.4 \\
& DoC ($\tau{=}0.8$) \cite{guillory2021predicting} & 15.5 $\pm$ 1.3 & 16.0 $\pm$ 1.3 & 17.3 $\pm$ 1.4 & 19.2 $\pm$ 1.6 & 20.5 $\pm$ 1.6 & 17.7 $\pm$ 1.4 \\
& DoC ($\tau{=}0.9$) \cite{guillory2021predicting} & 16.7 $\pm$ 1.4 & 17.3 $\pm$ 1.4 & 18.6 $\pm$ 1.5 & 20.3 $\pm$ 1.7 & 21.7 $\pm$ 1.7 & 18.9 $\pm$ 1.5 \\
& PseAutoEval \cite{ren2024autoeval} & 11.6 $\pm$ 0.9 & 12.2 $\pm$ 1.0 & 13.1 $\pm$ 1.0 & 15.1 $\pm$ 1.2 & 16.3 $\pm$ 1.3 & 13.7 $\pm$ 1.1 \\
& BugJudge \cite{liu2025nl2sqlbugs} & 14.8 $\pm$ 1.2 & 15.4 $\pm$ 1.2 & 16.2 $\pm$ 1.3 & 18.1 $\pm$ 1.4 & 19.0 $\pm$ 1.5 & 16.7 $\pm$ 1.3 \\
& ArenaCmp \cite{zheng2023llmasjudge} & 9.7 $\pm$ 0.8 & 10.4 $\pm$ 0.9 & 11.2 $\pm$ 0.9 & 12.6 $\pm$ 1.0 & 13.5 $\pm$ 1.1 & 11.5 $\pm$ 0.9 \\
& FusionSQL-TL & \underline{3.4 $\pm$ 1.2} & \underline{4.0 $\pm$ 1.2} & \underline{4.6 $\pm$ 1.3} & \underline{5.2 $\pm$ 1.4} & \underline{5.6 $\pm$ 1.4} & \underline{4.6 $\pm$ 1.3} \\
& FusionSQL (Ours) & \textbf{3.1 $\pm$ 0.5} & \textbf{3.7 $\pm$ 0.5} & \textbf{4.2 $\pm$ 0.6} & \textbf{4.8 $\pm$ 0.7} & \textbf{5.1 $\pm$ 0.7} & \textbf{4.2 $\pm$ 0.6} \\
\midrule
\multirow{10}{*}{WikiSQL $\rightarrow$ Spider}
& ATC-MC \cite{garg2022leveraging} & 12.2 $\pm$ 1.0 & 13.1 $\pm$ 1.1 & 13.8 $\pm$ 1.2 & 15.2 $\pm$ 1.3 & 16.1 $\pm$ 1.4 & 14.1 $\pm$ 1.2 \\
& ATC-NE \cite{garg2022leveraging} & 13.4 $\pm$ 1.1 & 14.0 $\pm$ 1.2 & 15.1 $\pm$ 1.3 & 16.3 $\pm$ 1.4 & 17.5 $\pm$ 1.5 & 15.3 $\pm$ 1.3 \\
& DoC ($\tau{=}0.8$) \cite{guillory2021predicting} & 14.6 $\pm$ 1.2 & 15.3 $\pm$ 1.3 & 16.5 $\pm$ 1.4 & 17.8 $\pm$ 1.5 & 19.0 $\pm$ 1.6 & 16.6 $\pm$ 1.4 \\
& DoC ($\tau{=}0.9$) \cite{guillory2021predicting} & 15.8 $\pm$ 1.3 & 16.4 $\pm$ 1.3 & 17.7 $\pm$ 1.4 & 19.1 $\pm$ 1.6 & 20.3 $\pm$ 1.6 & 17.9 $\pm$ 1.4 \\
& PseAutoEval \cite{ren2024autoeval} & 11.1 $\pm$ 0.9 & 11.8 $\pm$ 1.0 & 12.6 $\pm$ 1.0 & 13.7 $\pm$ 1.1 & 14.9 $\pm$ 1.2 & 12.8 $\pm$ 1.0 \\
& BugJudge \cite{liu2025nl2sqlbugs} & 13.6 $\pm$ 1.1 & 14.2 $\pm$ 1.1 & 15.1 $\pm$ 1.2 & 16.5 $\pm$ 1.3 & 17.6 $\pm$ 1.4 & 15.4 $\pm$ 1.2 \\
& ArenaCmp \cite{zheng2023llmasjudge} & 9.2 $\pm$ 0.8 & 9.9 $\pm$ 0.8 & 10.7 $\pm$ 0.9 & 12.0 $\pm$ 1.0 & 12.8 $\pm$ 1.0 & 10.9 $\pm$ 0.9 \\
& FusionSQL-TL & \underline{3.6 $\pm$ 1.2} & \underline{4.1 $\pm$ 1.2} & \underline{4.7 $\pm$ 1.3} & \underline{5.1 $\pm$ 1.3} & \underline{5.6 $\pm$ 1.4} & \underline{4.6 $\pm$ 1.3} \\
& FusionSQL (Ours) & \textbf{3.2 $\pm$ 0.5} & \textbf{3.8 $\pm$ 0.5} & \textbf{4.3 $\pm$ 0.6} & \textbf{4.7 $\pm$ 0.7} & \textbf{5.2 $\pm$ 0.7} & \textbf{4.2 $\pm$ 0.6} \\
\midrule
\multirow{10}{*}{SParC $\rightarrow$ CoSQL (in-domain)}
& ATC-MC \cite{garg2022leveraging} & 6.5 $\pm$ 0.6 & 7.2 $\pm$ 0.7 & 7.8 $\pm$ 0.8 & 8.3 $\pm$ 0.8 & 9.0 $\pm$ 0.9 & 7.8 $\pm$ 0.8 \\
& ATC-NE \cite{garg2022leveraging} & 7.1 $\pm$ 0.6 & 7.8 $\pm$ 0.7 & 8.4 $\pm$ 0.7 & 9.0 $\pm$ 0.8 & 9.6 $\pm$ 0.9 & 8.4 $\pm$ 0.7 \\
& DoC ($\tau{=}0.8$) \cite{guillory2021predicting} & 7.7 $\pm$ 0.6 & 8.3 $\pm$ 0.7 & 8.8 $\pm$ 0.7 & 9.3 $\pm$ 0.8 & 9.9 $\pm$ 0.8 & 8.8 $\pm$ 0.7 \\
& DoC ($\tau{=}0.9$) \cite{guillory2021predicting} & 8.8 $\pm$ 0.7 & 9.3 $\pm$ 0.7 & 9.8 $\pm$ 0.8 & 10.4 $\pm$ 0.9 & 10.9 $\pm$ 0.9 & 9.8 $\pm$ 0.8 \\
& PseAutoEval \cite{ren2024autoeval} & 5.5 $\pm$ 0.5 & 6.1 $\pm$ 0.5 & 6.7 $\pm$ 0.6 & 7.2 $\pm$ 0.6 & 7.8 $\pm$ 0.7 & 6.7 $\pm$ 0.6 \\
& BugJudge \cite{liu2025nl2sqlbugs} & 6.1 $\pm$ 0.6 & 6.7 $\pm$ 0.6 & 7.3 $\pm$ 0.7 & 7.9 $\pm$ 0.7 & 8.4 $\pm$ 0.8 & 7.3 $\pm$ 0.7 \\
& ArenaCmp \cite{zheng2023llmasjudge} & 3.9 $\pm$ 0.4 & 4.4 $\pm$ 0.4 & 4.9 $\pm$ 0.5 & 5.4 $\pm$ 0.5 & 5.9 $\pm$ 0.5 & 4.9 $\pm$ 0.5 \\
& FusionSQL-TL & \textbf{1.5 $\pm$ 1.2} & \textbf{1.7 $\pm$ 1.2} & \textbf{2.0 $\pm$ 1.3} & \textbf{2.2 $\pm$ 1.3} & \textbf{2.4 $\pm$ 1.3} & \textbf{2.0 $\pm$ 1.3} \\
& FusionSQL (Ours) & \underline{1.6 $\pm$ 0.3} & \underline{1.8 $\pm$ 0.3} & \underline{2.1 $\pm$ 0.3} & \underline{2.3 $\pm$ 0.4} & \underline{2.5 $\pm$ 0.4} & \underline{2.1 $\pm$ 0.3} \\
\midrule
\multirow{10}{*}{Spider $\rightarrow$ SynSQL-2.5M}
& ATC-MC \cite{garg2022leveraging}        & 10.9 $\pm$ 0.9 & 11.7 $\pm$ 1.0 & 12.3 $\pm$ 1.0 & 13.8 $\pm$ 1.1 & 14.7 $\pm$ 1.2 & 12.7 $\pm$ 1.0 \\
& ATC-NE \cite{garg2022leveraging}        & 12.1 $\pm$ 1.0 & 12.9 $\pm$ 1.1 & 13.5 $\pm$ 1.1 & 14.9 $\pm$ 1.2 & 15.8 $\pm$ 1.3 & 13.8 $\pm$ 1.1 \\
& DoC ($\tau{=}0.8$) \cite{guillory2021predicting} & 12.9 $\pm$ 1.0 & 13.6 $\pm$ 1.1 & 14.7 $\pm$ 1.2 & 16.0 $\pm$ 1.3 & 17.2 $\pm$ 1.4 & 14.9 $\pm$ 1.2 \\
& DoC ($\tau{=}0.9$) \cite{guillory2021predicting} & 14.1 $\pm$ 1.1 & 14.8 $\pm$ 1.2 & 15.9 $\pm$ 1.3 & 17.2 $\pm$ 1.4 & 18.4 $\pm$ 1.5 & 16.1 $\pm$ 1.3 \\
& PseAutoEval \cite{ren2024autoeval}     & 9.5 $\pm$ 0.8 & 10.1 $\pm$ 0.9 & 10.8 $\pm$ 0.9 & 12.0 $\pm$ 1.0 & 13.1 $\pm$ 1.1 & 11.1 $\pm$ 0.9 \\
& BugJudge \cite{liu2025nl2sqlbugs} & 12.4 $\pm$ 1.0 & 13.2 $\pm$ 1.1 & 14.0 $\pm$ 1.1 & 15.5 $\pm$ 1.2 & 16.6 $\pm$ 1.3 & 14.3 $\pm$ 1.1 \\
& ArenaCmp \cite{zheng2023llmasjudge} & 8.4 $\pm$ 0.7 & 9.1 $\pm$ 0.8 & 9.8 $\pm$ 0.8 & 11.1 $\pm$ 0.9 & 11.9 $\pm$ 1.0 & 10.1 $\pm$ 0.8 \\
& FusionSQL-TL & \underline{3.1 $\pm$ 1.2} & \underline{3.5 $\pm$ 1.2} & \underline{4.0 $\pm$ 1.3} & \underline{4.4 $\pm$ 1.3} & \underline{4.9 $\pm$ 1.4} & \underline{4.0 $\pm$ 1.3} \\
& FusionSQL (Ours)           & \textbf{2.8 $\pm$ 0.4} & \textbf{3.2 $\pm$ 0.5} & \textbf{3.7 $\pm$ 0.5} & \textbf{4.1 $\pm$ 0.6} & \textbf{4.5 $\pm$ 0.6} & \textbf{3.7 $\pm$ 0.5} \\
\midrule
\multirow{10}{*}{WikiSQL $\rightarrow$ Spider 2.0}
& ATC-MC \cite{garg2022leveraging}        & 18.0 $\pm$ 1.5 & 18.7 $\pm$ 1.5 & 19.6 $\pm$ 1.6 & 21.0 $\pm$ 1.7 & 22.2 $\pm$ 1.8 & 19.9 $\pm$ 1.6 \\
& ATC-NE \cite{garg2022leveraging}        & 19.4 $\pm$ 1.6 & 20.1 $\pm$ 1.7 & 21.3 $\pm$ 1.8 & 22.6 $\pm$ 1.9 & 23.9 $\pm$ 2.0 & 21.5 $\pm$ 1.8 \\
& DoC ($\tau{=}0.8$) \cite{guillory2021predicting} & 20.5 $\pm$ 1.7 & 21.3 $\pm$ 1.8 & 22.7 $\pm$ 1.9 & 24.0 $\pm$ 2.0 & 25.4 $\pm$ 2.1 & 22.8 $\pm$ 1.9 \\
& DoC ($\tau{=}0.9$) \cite{guillory2021predicting} & 21.7 $\pm$ 1.8 & 22.5 $\pm$ 1.9 & 23.9 $\pm$ 2.0 & 25.2 $\pm$ 2.1 & 26.6 $\pm$ 2.2 & 23.9 $\pm$ 2.0 \\
& PseAutoEval \cite{ren2024autoeval}     & 16.3 $\pm$ 1.3 & 17.0 $\pm$ 1.4 & 17.7 $\pm$ 1.4 & 18.8 $\pm$ 1.5 & 20.1 $\pm$ 1.6 & 18.0 $\pm$ 1.4 \\
& BugJudge \cite{liu2025nl2sqlbugs} & 17.3 $\pm$ 1.4 & 18.1 $\pm$ 1.5 & 19.3 $\pm$ 1.6 & 20.7 $\pm$ 1.7 & 22.0 $\pm$ 1.8 & 19.5 $\pm$ 1.6 \\
& ArenaCmp \cite{zheng2023llmasjudge} & 12.6 $\pm$ 1.0 & 13.4 $\pm$ 1.1 & 14.5 $\pm$ 1.2 & 15.8 $\pm$ 1.3 & 16.9 $\pm$ 1.4 & 14.6 $\pm$ 1.2 \\
& FusionSQL-TL & \underline{4.5 $\pm$ 1.3} & \underline{5.1 $\pm$ 1.4} & \underline{5.6 $\pm$ 1.4} & \underline{6.1 $\pm$ 1.5} & \underline{6.6 $\pm$ 1.5} & \underline{5.6 $\pm$ 1.4} \\
& FusionSQL (Ours)           & \textbf{4.2 $\pm$ 0.6} & \textbf{4.8 $\pm$ 0.7} & \textbf{5.3 $\pm$ 0.7} & \textbf{5.8 $\pm$ 0.8} & \textbf{6.3 $\pm$ 0.8} & \textbf{5.3 $\pm$ 0.7} \\
\bottomrule
\end{tabular}
}
\vspace{-1em}
\end{table*}

\sstitle{Budget Constraints}
FusionDataset is generated under a fixed budget $B=1{,}000$ (USD), so we formulate spending at the database level and constrain it explicitly. Let $\mathcal{D}$ denote the set of databases, and for each $d \in \mathcal{D}$ let $n^{\text{gen}}_d$, $n^{\text{val}}_d$, and $n^{\text{exec}}_d$ be the number of generated candidates, lightweight validation checks (schema/type checks, parsing, and canonicalization), and execution checks, with effective per-sample costs $c^{\text{gen}}$, $c^{\text{val}}$, and $c^{\text{exec}}$ (amortized with batched prompts), respectively. The total cost is:
\begin{equation}
    C = \sum_{d \in \mathcal{D}} \big(n^{\text{gen}}_d c^{\text{gen}} + n^{\text{val}}_d c^{\text{val}} + n^{\text{exec}}_d c^{\text{exec}}\big) \le B.
    \label{eq:budget_constraint}
\end{equation}
The per-sample costs are $c^{\text{gen}}=0.00012$, $c^{\text{val}}=0.00003$, and $c^{\text{exec}}=0.0004$, and we budget for expected totals $\sum_{d \in \mathcal{D}} n^{\text{gen}}_d = 1.05N$, $\sum_{d \in \mathcal{D}} n^{\text{val}}_d = 1.05N$, and $\sum_{d \in \mathcal{D}} n^{\text{exec}}_d = 0.10N$ over $N=3{,}373{,}204$ accepted pairs, yielding $C \approx N(1.05(c^{\text{gen}}+c^{\text{val}})+0.10c^{\text{exec}})=666.21 \le B$. We enforce per-database caps by setting $\bar{n}^{\text{gen}}=160$ and $\bar{n}^{\text{exec}}=40$ as the maximum allowed generation and execution checks per database, and require $n^{\text{gen}}_d \le \bar{n}^{\text{gen}}$ and $n^{\text{exec}}_d \le \bar{n}^{\text{exec}}$. These constraints are chosen to satisfy the worst-case bound $|\mathcal{D}|(\bar{n}^{\text{gen}}(c^{\text{gen}}+c^{\text{val}})+\bar{n}^{\text{exec}}c^{\text{exec}})=985.00 \le B$. Sampling proceeds in batches of size $b=24$. These controls keep costly execution checks bounded and lightweight checks to filter most candidates. Caching schema summaries and reusing validated SQL for paraphrase generation further amortizes $n^{\text{gen}}_d$, ensuring $C$ remains within budget while still yielding 3.37M pairs and 3.12M unique SQL.

\sstitle{Training Data Formation} From this corpus, we construct training data by forming a \textit{meta-set} of instances $(\mathcal{D}_{\mathrm{train}}, S_i)$, where each \textit{sample set} $S_i$ is independently drawn from FusionDataset to induce a distinct train--test shift. We generate a meta-set of $n{=}30$K such sample sets, with $|S_i|\!\leq\!10$K, spanning a wide range of schema structures, query patterns, and linguistic variations. $n$ and $|S_i|$ balance coverage and costs, which is proven in subsequent analysis.

\sstitle{Sample-Set Size Effect} We analyze the impact of training data sources by training FusionSQL on meta-instances drawn from different candidate datasets while keeping the evaluator architecture and transfer settings fixed. Using FusionDataset (3.3M samples), we construct 30K training instances with $|S_i|\!\leq\!10$K, compute shift descriptors $\Delta_{\text{train}}$, and train FusionSQL to predict EX. At inference, accuracy on an unseen, unlabeled target is estimated via $\Delta_{\text{test}}$. We repeat the same procedure with alternative sources (e.g., Spider, BIRD) under identical cross-dataset transfers. As shown in \autoref{fig:fusiondataset_samplesize}, MAE (avg. across five transfers in \autoref{tab:mae-evaluator}) decreases as sample sets grow for all sources, but performance is ultimately determined by representativeness rather than scale: WikiSQL still exhibits higher error than FusionDataset at matched sizes, whereas FusionDataset achieves low MAE even with small samples (e.g., $\approx6$ at $2$K). These results confirm that broad coverage is the primary driver of reliable label-free accuracy estimation, rather than raw volume.

\begin{figure}[!h]
  \vspace{-1em}
  \centering
  \includegraphics[width=0.9\linewidth]{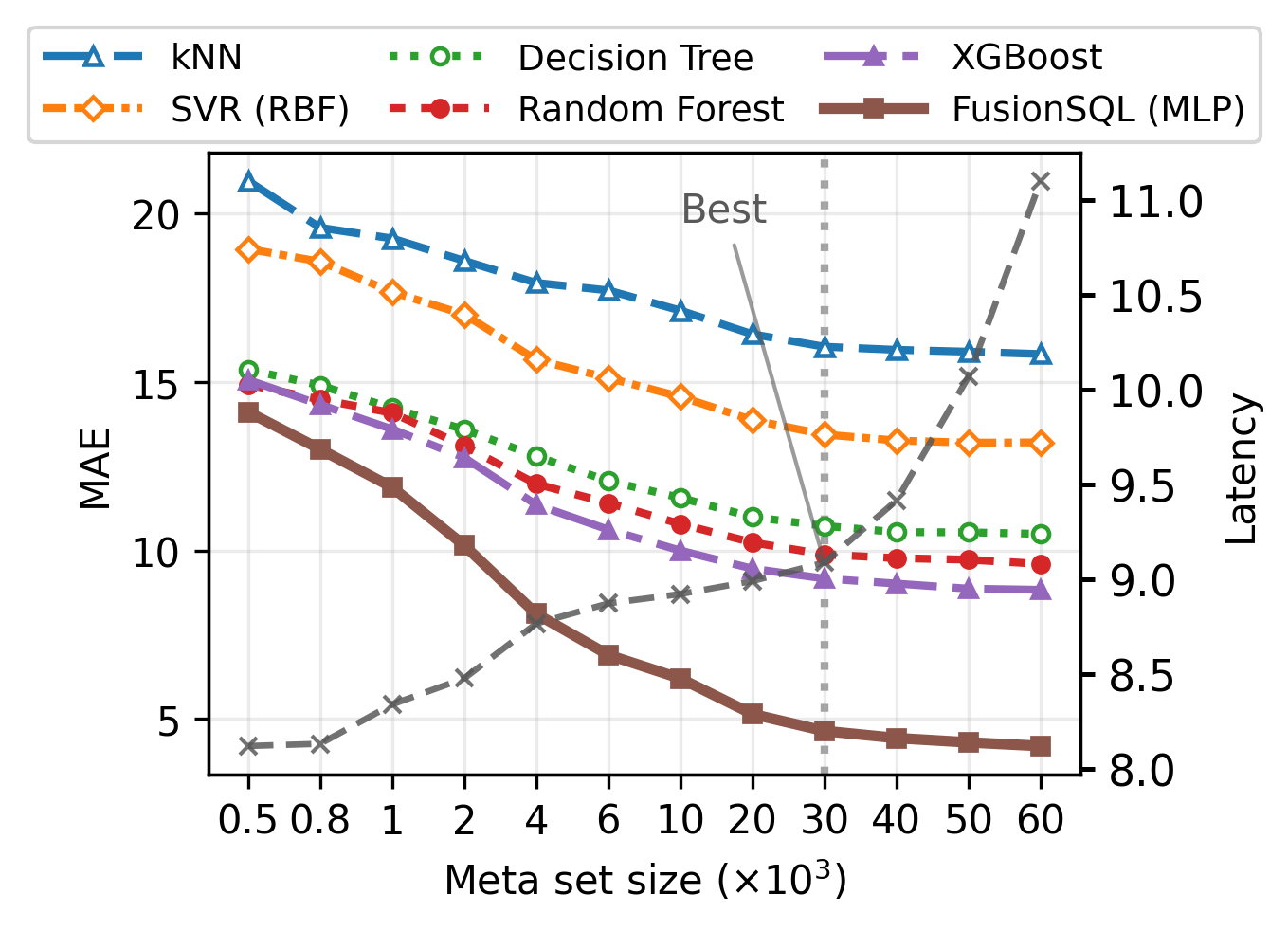}
  \vspace{-.5em}
  \caption{\textbf{Meta-set size.} FusionSQL's MLP attains the lowest error and benefits most from larger meta-sets, while costs rise sharply beyond $30{,}000$ with marginal gains.}
  \label{fig:meta_set_size}
  \vspace{-1em}
\end{figure}

\sstitle{Meta-Set Size Effect}
To understand the trade--offs between model capacity and training data availability, we jointly study how the choice of regression architecture and the meta-set size $n$ affect estimation performance. Each meta instance $(\mathcal{D}_{\mathrm{train}}, S_i)$, with $S_i$ sampled from FusionDataset, encodes a distinct train--test shift through $\Delta_{\text{train}}=h(\phi_{\mathcal{D}_{\mathrm{train}}},\phi_{S_i})$. We compare FusionSQL's MLP against classical regressors (kNN, SVR, Decision Trees, Random Forests, and XGBoost) while varying $n$. As illustrated in \autoref{fig:meta_set_size}, estimation error decreases with larger meta-sets for all methods, but the MLP benefits most and continues to improve up to $n{=}30$K, whereas simpler models saturate earlier. Training cost increases gradually with $n$ but rises sharply beyond $n{=}30$K, after which performance gains become marginal relative to computational overhead. We therefore adopt an MLP and fix $n{=}30$K for all experiments.


Overall, results show that FusionSQL can achieve reliable accuracy estimation without requiring exhaustive training data. Under realistic constraints on time and cost, practitioners can select the size of sample set and meta-set according to the target MAE required for deployment. This enables organizations to scale the meta-set only when tighter error bounds are necessary.

\subsection{Evaluator Learning}
\label{sec:evaluator_learning_experiment}

To address \textbf{RQ3 (Evaluator Learning)}, we benchmark FusionSQL's ability to estimate accuracy on unseen and unlabeled Text2SQL datasets, and evaluate how well this transfers across unseen dataset distributions and Text2SQL model families.

\begin{figure}[h]
\vspace{-1em}
    \centering
    \includegraphics[width=1\linewidth]{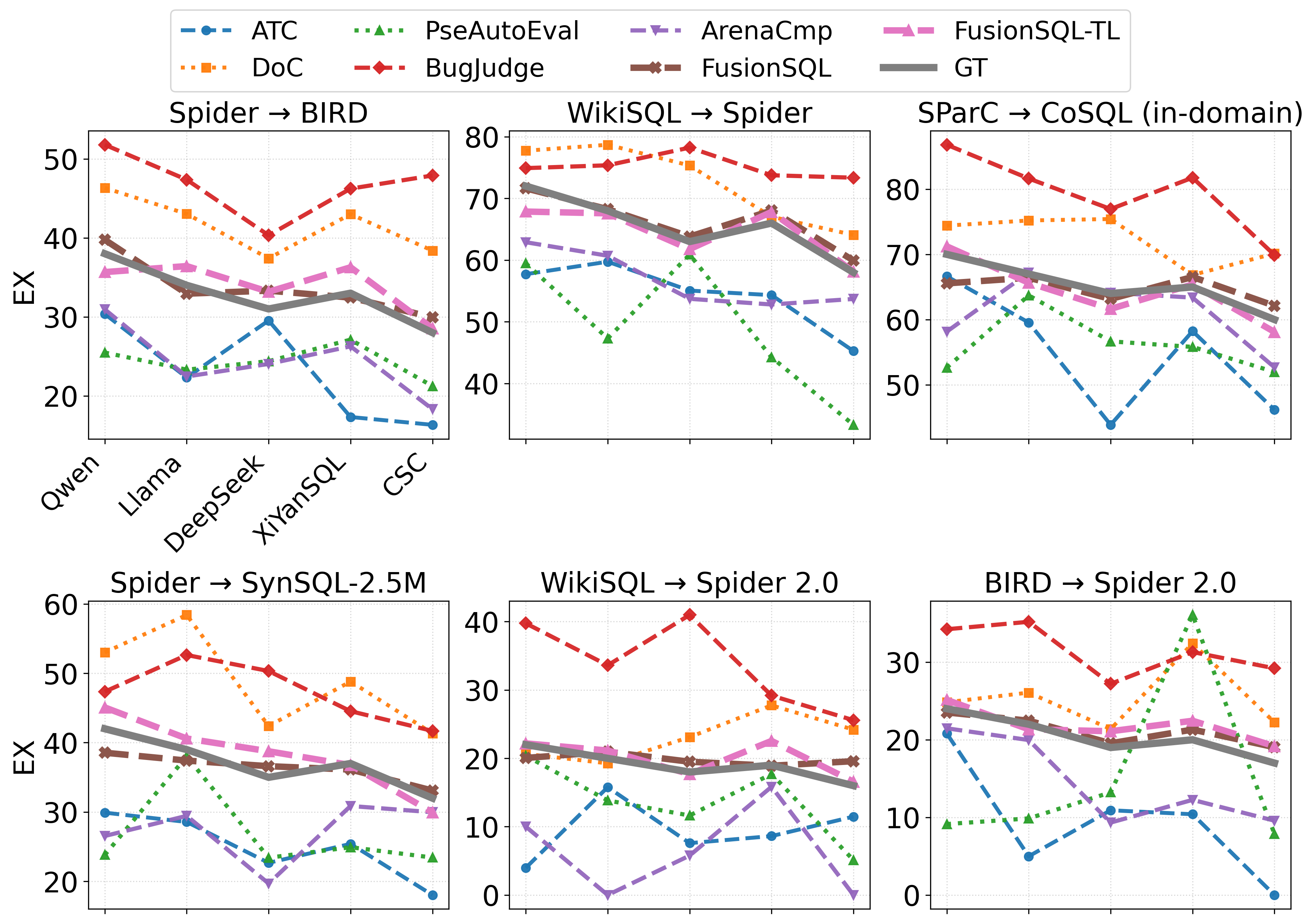}
    \vspace{-0.5em}
    \caption{\textbf{EX across transfers.} Across six source--target transfers and five base models, FusionSQL closely follows the ground-truth (GT) trend, while FusionSQL-TL exhibits mild bias on harder targets due to limited labeled adaptation.}
    \label{fig:fusiondataset_estimation_calibration}
    \vspace{-1em}
\end{figure}

\sstitle{Evaluator Benchmark}
As this setting has not been directly addressed before, we adapt representative label-free baselines, including confidence-based methods ATC, including its two variants ATC-MC (maximum confidence) and ATC-NE (negative entropy) \cite{garg2022leveraging}, DoC \cite{guillory2021predicting}, and PseAutoEval \cite{ren2024autoeval}. We also include two LLM-as-a-judge baselines, BugJudge \cite{liu2025nl2sqlbugs} and ArenaCmp \cite{zheng2023llmasjudge}, both using GPT-5 as the judge. FusionSQL is trained purely from shift descriptors between the source training set and FusionDataset samples, reflecting a zero-label transfer setting. We additionally report a transfer learning approach, FusionSQL-TL, which fine-tunes the evaluator with a small labeled target subset, with $p\!\in\!\{10,25\}\%$ being the proportion of labeled examples, to test whether limited supervision improves estimation. We evaluate five source--target transfers and five Text2SQL models, measuring MAE (percentage points) between predicted and true EX, averaged over five seeds with 95\% confidence intervals. As shown in \autoref{tab:mae-evaluator}, FusionSQL achieves the lowest MAE in most cases with tight confidence bounds, while FusionSQL-TL (average of $p\!=\!10\%$ and $p\!=\!25\%$) is occasionally competitive but suffers higher variance and bias under large shifts due to limited labeled coverage. Overall, FusionSQL generalizes robustly across unseen workloads, enabling reliable label-free accuracy estimation under severe distribution shift.

\begin{figure}[h]
  \vspace{-1em}
    \centering
    \includegraphics[width=0.8\linewidth]{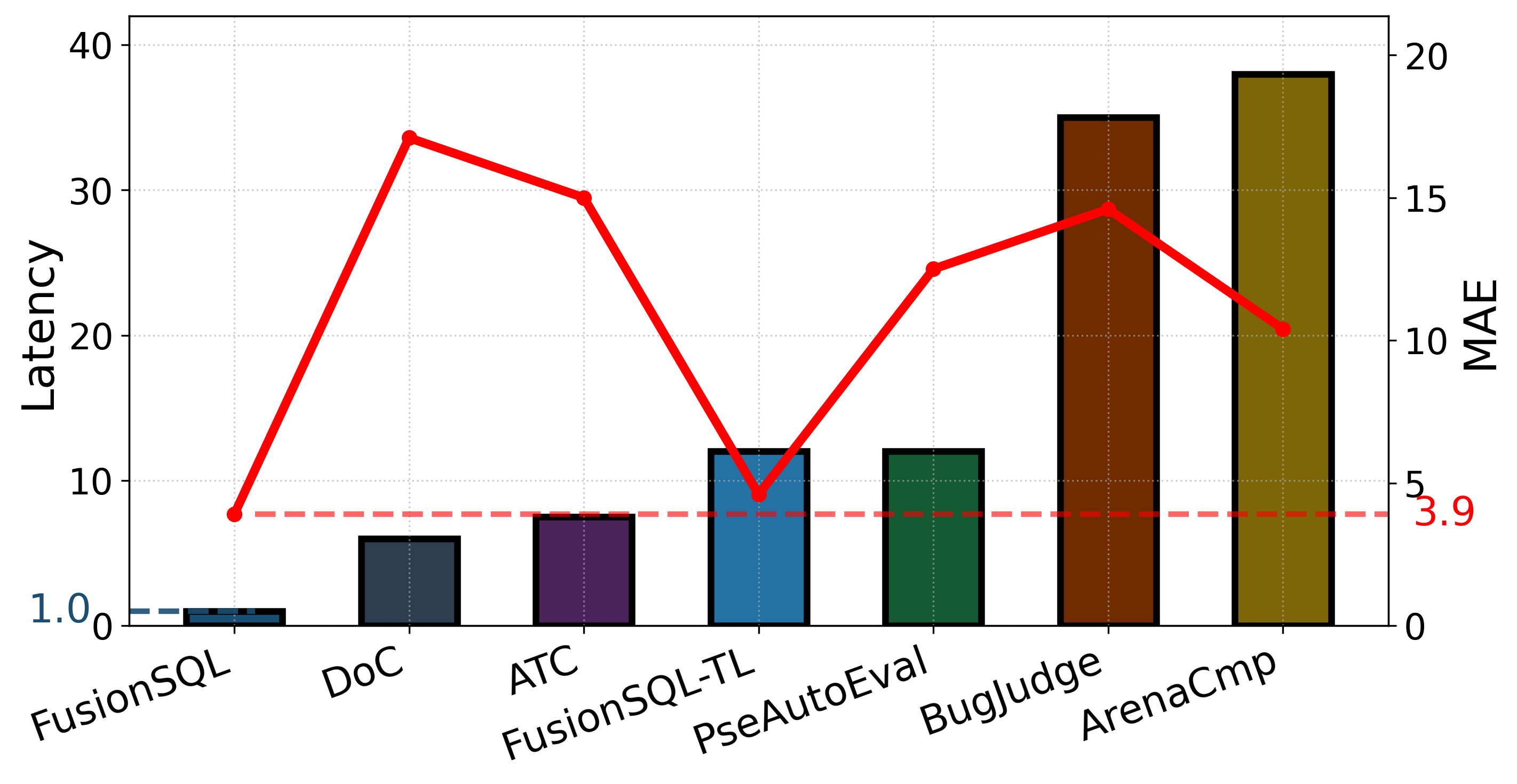}
    \vspace{-0.5em}
    \caption{\textbf{Latency--accuracy trade-off.} Bars report average evaluation latency, while the red curve shows MAE (lower is better). FusionSQL offers the best trade-off, whereas judge-based methods incur high latency without much MAE gains.}
    \label{fig:latency-vs-mae}
    \vspace{-1em}
\end{figure}

\sstitle{Estimation Calibration} 
MAE remains the primary evaluation metric because it directly measures the error between estimated and gold EX. However, EX curves reveal whether an estimator is biased toward over- or under-estimation. In \autoref{fig:fusiondataset_estimation_calibration}, FusionSQL closely tracks the GT line across models, indicating good calibration. On simpler targets (e.g., WikiSQL $\rightarrow$ Spider and SParC $\rightarrow$ CoSQL), FusionSQL slightly underestimates performance, reflecting conservative estimates when schema overlap is high. On harder targets (BIRD, Spider~2.0, SynSQL-2.5M), FusionSQL is occasionally above GT, but the deviations are limited. In contrast, FusionSQL-TL tends to overestimate performance on complex targets because the small labeled subset used for transfer learning is biased toward frequent patterns and does not adequately cover tail queries. As a result, the added supervision introduces a misleading adaptation signal that worsens performance rather than improving it, since FusionDataset already provides broad coverage of schema structures and query patterns.

\begin{figure}[!h]
    \centering
    \includegraphics[width=0.9\linewidth]{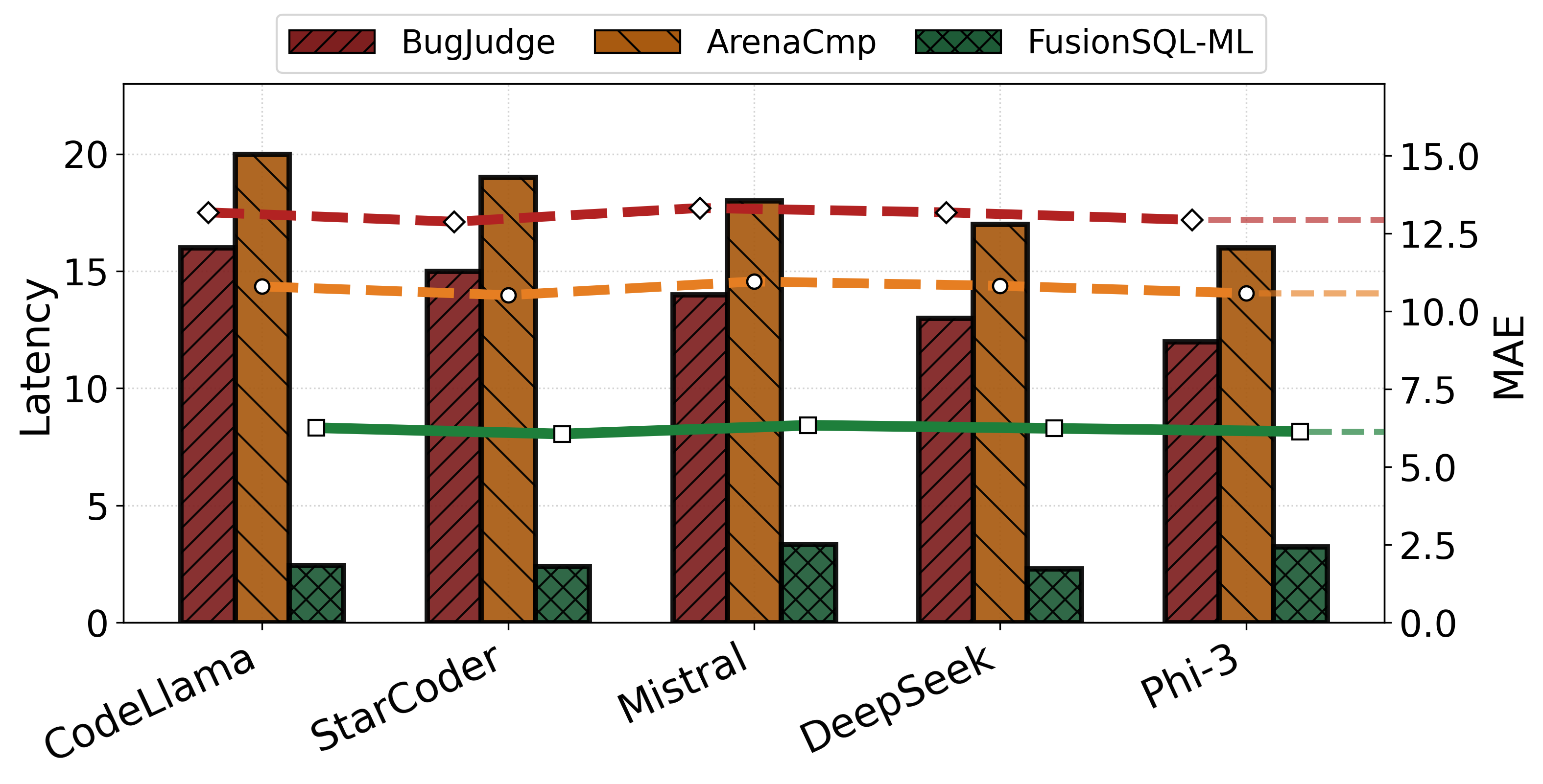}
    \caption{\textbf{Latency--accuracy trade-off on unseen Text2SQL models.} Bars denote relative latency, while lines show MAE. FusionSQL-ML attains the lowest MAE with low latency.}
    \label{fig:unseen-latency-mae}
    \vspace{-1em}
\end{figure}

\begin{table*}[h]
\centering
\footnotesize
\setlength{\tabcolsep}{3.6pt}
\caption{MAE ($\downarrow$) for generalizing FusionSQL to new Text2SQL models (mean $\pm$ 95\% CI, percentage points). Columns are the unseen model pool. Best in \textbf{bold}.}
\label{tab:mae-unseen-models}
\resizebox{0.85\textwidth}{!}{%
\begin{tabular}{llcccccc}
\toprule
\textbf{Transfer ($A \rightarrow B$)} &
\textbf{Methods} &
\textit{CodeLlama-34B} &
\textit{StarCoder2-15B} &
\textit{Mistral-7B} &
\textit{DeepSeek-Coder-6.7B} &
\textit{Phi-3-mini} &
\textit{Avg.} \\
\midrule
{Spider $\rightarrow$ BIRD}
& BugJudge \cite{liu2025nl2sqlbugs} & 13.8 $\pm$ 1.0 & 13.5 $\pm$ 1.1 & 14.0 $\pm$ 1.0 & 13.9 $\pm$ 0.9 & 13.6 $\pm$ 1.0 & 13.8 $\pm$ 1.0 \\
& ArenaCmp \cite{zheng2023llmasjudge} & 11.1 $\pm$ 0.8 & 10.8 $\pm$ 0.9 & 11.4 $\pm$ 1.0 & 11.2 $\pm$ 0.9 & 10.9 $\pm$ 0.8 & 11.1 $\pm$ 0.9 \\
& FusionSQL-ML (Ours) & \textbf{6.7 $\pm$ 0.5} & \textbf{6.5 $\pm$ 0.6} & \textbf{6.8 $\pm$ 0.7} & \textbf{6.7 $\pm$ 0.6} & \textbf{6.6 $\pm$ 0.5} & \textbf{6.7 $\pm$ 0.6} \\
\midrule
{WikiSQL $\rightarrow$ Spider}
& BugJudge \cite{liu2025nl2sqlbugs} & 12.7 $\pm$ 1.0 & 12.4 $\pm$ 1.1 & 12.9 $\pm$ 1.0 & 12.8 $\pm$ 0.9 & 12.5 $\pm$ 1.0 & 12.7 $\pm$ 1.0 \\
& ArenaCmp \cite{zheng2023llmasjudge} & 10.4 $\pm$ 0.8 & 10.1 $\pm$ 0.9 & 10.6 $\pm$ 1.0 & 10.4 $\pm$ 0.9 & 10.2 $\pm$ 0.8 & 10.3 $\pm$ 0.9 \\
& FusionSQL-ML (Ours) & \textbf{6.0 $\pm$ 0.4} & \textbf{5.8 $\pm$ 0.5} & \textbf{6.1 $\pm$ 0.6} & \textbf{6.0 $\pm$ 0.5} & \textbf{5.9 $\pm$ 0.4} & \textbf{6.0 $\pm$ 0.5} \\
\midrule
{SParC $\rightarrow$ CoSQL}
& BugJudge \cite{liu2025nl2sqlbugs} & 11.5 $\pm$ 0.8 & 11.3 $\pm$ 0.9 & 11.6 $\pm$ 1.0 & 11.5 $\pm$ 0.9 & 11.2 $\pm$ 0.8 & 11.4 $\pm$ 0.9 \\
& ArenaCmp \cite{zheng2023llmasjudge} & 9.6 $\pm$ 0.7 & 9.4 $\pm$ 0.8 & 9.7 $\pm$ 0.9 & 9.6 $\pm$ 0.8 & 9.3 $\pm$ 0.7 & 9.5 $\pm$ 0.8 \\
& FusionSQL-ML (Ours) & \textbf{5.1 $\pm$ 0.4} & \textbf{4.9 $\pm$ 0.5} & \textbf{5.1 $\pm$ 0.6} & \textbf{5.0 $\pm$ 0.5} & \textbf{4.9 $\pm$ 0.4} & \textbf{5.0 $\pm$ 0.5} \\
\midrule
{Spider $\rightarrow$ SynSQL-2.5M}
& BugJudge \cite{liu2025nl2sqlbugs} & 13.3 $\pm$ 1.0 & 13.0 $\pm$ 1.1 & 13.4 $\pm$ 1.0 & 13.2 $\pm$ 0.9 & 13.1 $\pm$ 1.0 & 13.2 $\pm$ 1.0 \\
& ArenaCmp \cite{zheng2023llmasjudge} & 10.9 $\pm$ 0.8 & 10.6 $\pm$ 0.9 & 11.0 $\pm$ 1.0 & 10.9 $\pm$ 0.9 & 10.7 $\pm$ 0.8 & 10.8 $\pm$ 0.9 \\
& FusionSQL-ML (Ours) & \textbf{6.5 $\pm$ 0.5} & \textbf{6.3 $\pm$ 0.6} & \textbf{6.6 $\pm$ 0.7} & \textbf{6.5 $\pm$ 0.6} & \textbf{6.4 $\pm$ 0.5} & \textbf{6.5 $\pm$ 0.6} \\
\midrule
{WikiSQL $\rightarrow$ Spider~2.0}
& BugJudge \cite{liu2025nl2sqlbugs} & 14.6 $\pm$ 1.0 & 14.2 $\pm$ 1.1 & 14.7 $\pm$ 1.2 & 14.5 $\pm$ 1.1 & 14.3 $\pm$ 1.0 & 14.5 $\pm$ 1.1 \\
& ArenaCmp \cite{zheng2023llmasjudge} & 12.0 $\pm$ 0.9 & 11.7 $\pm$ 1.0 & 12.1 $\pm$ 1.1 & 12.0 $\pm$ 1.0 & 11.8 $\pm$ 0.9 & 11.9 $\pm$ 1.0 \\
& FusionSQL-ML (Ours) & \textbf{7.0 $\pm$ 0.5} & \textbf{6.8 $\pm$ 0.6} & \textbf{7.1 $\pm$ 0.7} & \textbf{7.0 $\pm$ 0.6} & \textbf{6.9 $\pm$ 0.5} & \textbf{7.0 $\pm$ 0.6} \\
\bottomrule
\end{tabular}
}
\end{table*}

\begin{figure*}[h]
  \centering
  \begin{subfigure}[t]{0.49\linewidth}
    \centering
    \includegraphics[width=\linewidth]{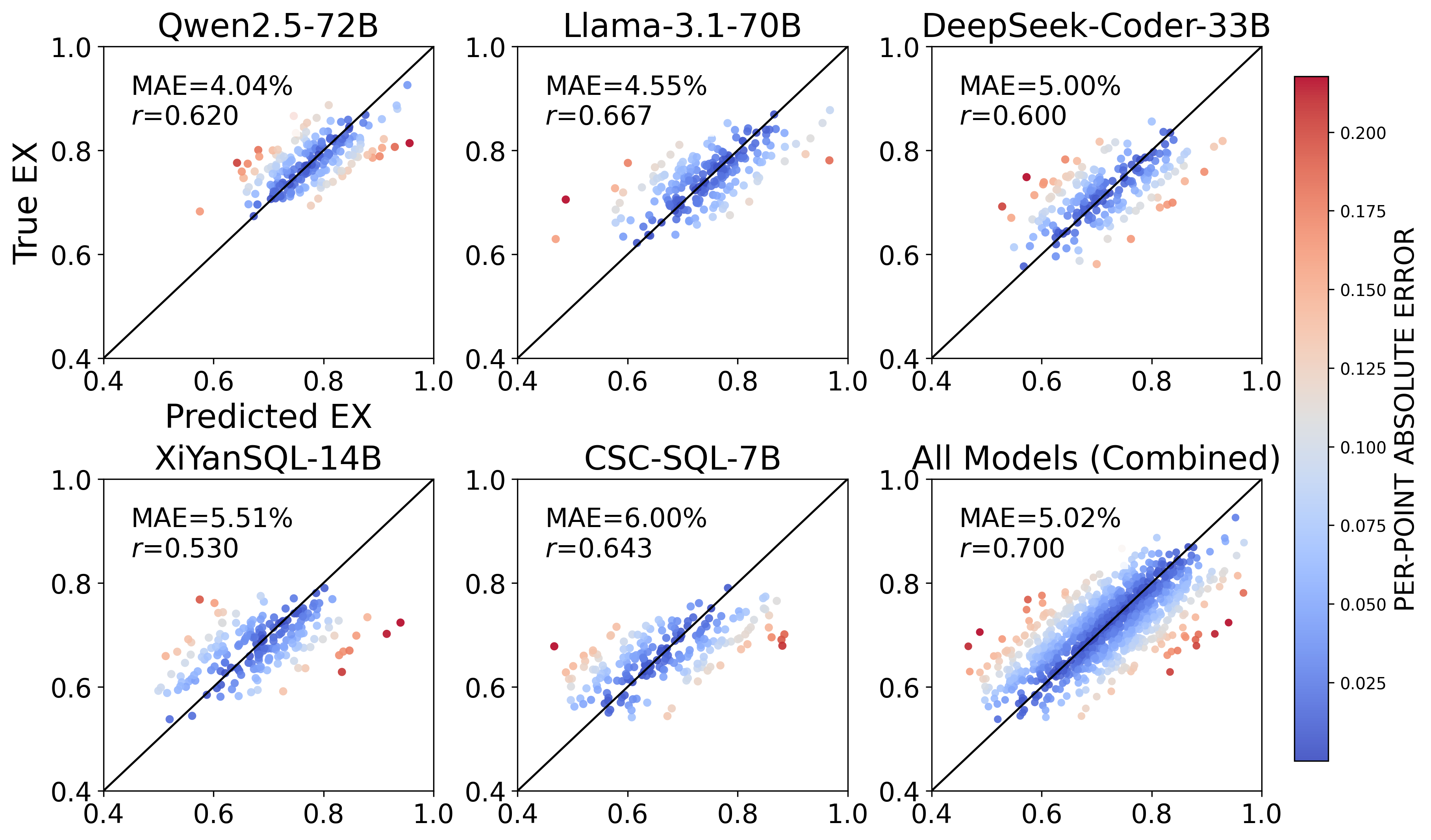}
    \caption{\textbf{Evaluator fidelity.} Predicted vs.\ true EX across Text2SQL models shows strong correlation and low MAE, indicating that \textit{FusionSQL} provides reliable dataset-level estimates without labels.}
    \label{fig:pred-vs-gold}
  \end{subfigure}
  \hfill
  \begin{subfigure}[t]{0.49\linewidth}
    \centering
    \includegraphics[width=\linewidth]{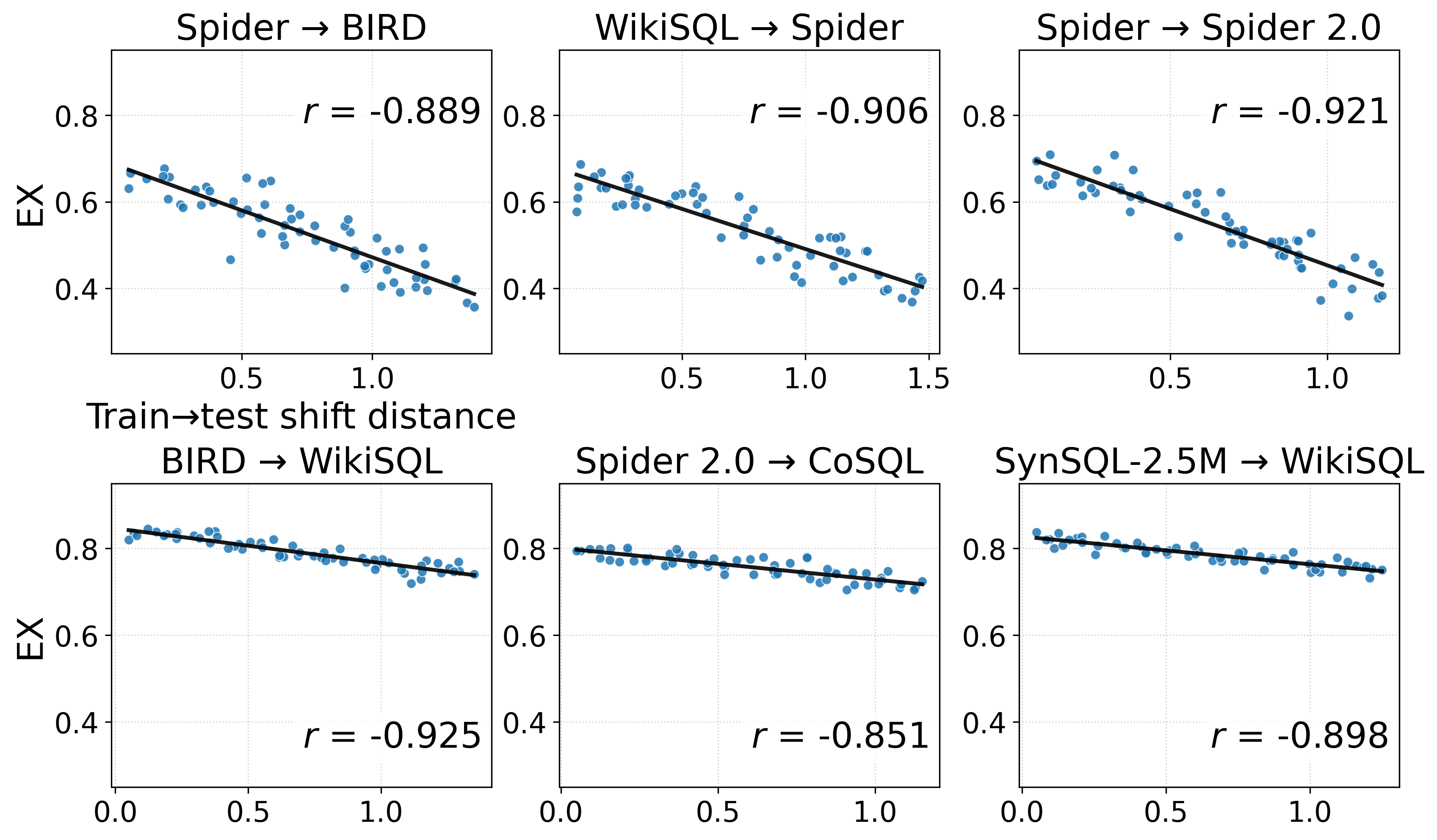}
    \caption{\textbf{Shift vs.\ EX.}
    Larger train--test shift generally lead to lower execution accuracy, while \textit{distant-but-easy} transfers exhibit only mild degradation, indicating that FusionSQL captures distributional mismatch beyond dataset difficulty.}
    \label{fig:shift-vs-accuracy}
  \end{subfigure}
  \caption{\textbf{FusionSQL reliability and shift sensitivity.} (a) FusionSQL accurately predicts EX. (b) Shift impacts EX on hard targets but only mildly on distant-but-easy transfers, demonstrating robustness.}
  \label{fig:fusionsql-evaluator}
  \vspace{-1em}
\end{figure*}

\sstitle{Latency} 
\autoref{fig:latency-vs-mae} summarizes per-target runtime (minutes, averaged over 5 transfers in \autoref{tab:mae-evaluator}) and highlights a clear speed--accuracy trade-off. FusionSQL is the fastest since it only computes a pooled dataset embedding from last-layer representations and applies a lightweight regressor. DoC and ATC are slower because they rely on additional per-example decoding and confidence statistics beyond the base Text2SQL generation. FusionSQL-TL and PseAutoEval incur extra latency from acquiring a small set of human labels for transfer-style calibration. FusionSQL-TL remains less accurate than FusionSQL due to bias and limited coverage from few labeled samples. Judge-based methods are the slowest: BugJudge and ArenaCmp require an additional LLM pass per example, resulting in substantially higher latency while still underperforming FusionSQL.

\sstitle{Evaluator Transfer to Unseen Models}
To extend the answer to \textbf{RQ2}, we evaluate whether FusionSQL generalizes to \textit{unseen} Text2SQL models using the FusionSQL-ML (FusionSQL meta-learning) described in \autoref{sec:further_consideration}. We meta-train a diverse pool of base models (20 total): Qwen2.5-72B/32B/14B/7B/3B/1.5B \cite{yang2025qwen3}, Llama-3.1-70B/8B \cite{grattafiori2024llama}, Mixtral-8x7B \cite{jiang2024mixtral}, DeepSeek-33B \cite{guo2024deepseek}, StarCoder2-7B/3B \cite{lozhkov2024starcoder}, CodeLlama-13B/7B \cite{roziere2023code}, Phi-3-medium \cite{abdin2024phi}, Yi-34B \cite{young2024yi}, Gemma-7B \cite{team2024gemma}, and XiYanSQL-14B \cite{XiYanSQL}, CSC-SQL-7B \cite{sheng2025csc}, InternLM-20B \cite{cai2024internlm2}. This pool intentionally excludes the unseen model pool in \autoref{tab:mae-unseen-models}. We compare against two LLM-as-a-judge baselines, BugJudge and ArenaCmp (both using GPT-5), which can evaluate any new LLM without retraining but incur substantial inference cost\footnote{See \href{https://platform.openai.com/docs/pricing}{OpenAI API pricing} for per-token costs.} and latency. As shown in \autoref{tab:mae-unseen-models}, FusionSQL-ML achieves the lowest MAE (5--7 points) across all unseen models and transfers, while ArenaCmp is consistently second best. BugJudge performs worse because error-detection approaches focus mainly on identifying obvious failures and may overlook false negatives, where a SQL query appears correct but still yields incorrect execution results. \autoref{fig:unseen-latency-mae} reports per-sample inference latency in minutes (bars) and MAE (lines): FusionSQL-ML remains both the most accurate and the fastest because it only requires pooled embeddings plus a lightweight evaluator, whereas judge-based methods require autoregressive generation and a separate judging pass for each example. Results confirm that meta-trained FusionSQL scales to new model families with strong accuracy and low latency.

\sstitle{Evaluator Fidelity and Shift Sensitivity}
To address \textbf{RQ3}, \autoref{fig:pred-vs-gold} and \autoref{fig:shift-vs-accuracy} jointly evaluate FusionSQL's reliability as a label-free evaluator across models and domains. \autoref{fig:pred-vs-gold} shows strong predictive fidelity: predicted EX closely matches gold EX with Pearson correlations $r{=}0.53$--$0.70$ and MAE below 6\%, indicating good calibration across diverse architectures and SQL generation behaviors, with only minor localized noise. \autoref{fig:shift-vs-accuracy} analyzes distance-sensitivity for Qwen2.5-72B using distance between the pooled embeddings of training and testing-set. Hard transfers show strong negative correlations with execution accuracy (e.g., Spider$\rightarrow$BIRD $r{=}-0.889$, WikiSQL$\rightarrow$Spider $r{=}-0.906$), whereas distant-but-easy transfers exhibit much flatter trends, with EX remaining high. This behavior is consistent with the broad schema and query diversity of FusionDataset, which exposes FusionSQL to representative shift patterns during training and keeps target distributions well supported despite large distances. This trend is reinforced by \autoref{fig:fusiondataset_samplesize}, where training on FusionDataset consistently achieves lower MAE across all sample budgets than on individual benchmarks (e.g., Spider, SParC, or SynSQL-2.5M), whose evaluators remain less accurate even at large sample sizes. Together, these results highlight FusionDataset's central role in FusionSQL's accuracy across both easy and hard transfer regimes.

\subsection{Ablation Study}
\label{sec:ablation}

We next conduct experiments to investigate \textbf{RQ4 (Scalability)} and \textbf{RQ5 (Shift Modeling)}. In particular, we study scalability in terms of memory footprint and latency, assess the applicability of FusionSQL to non-neural Text2SQL systems, and evaluate the robustness of different shift descriptors under diverse distributional shifts.

\begin{figure}[h]
  \vspace{-1em}
  \centering
  \includegraphics[width=0.8\linewidth]{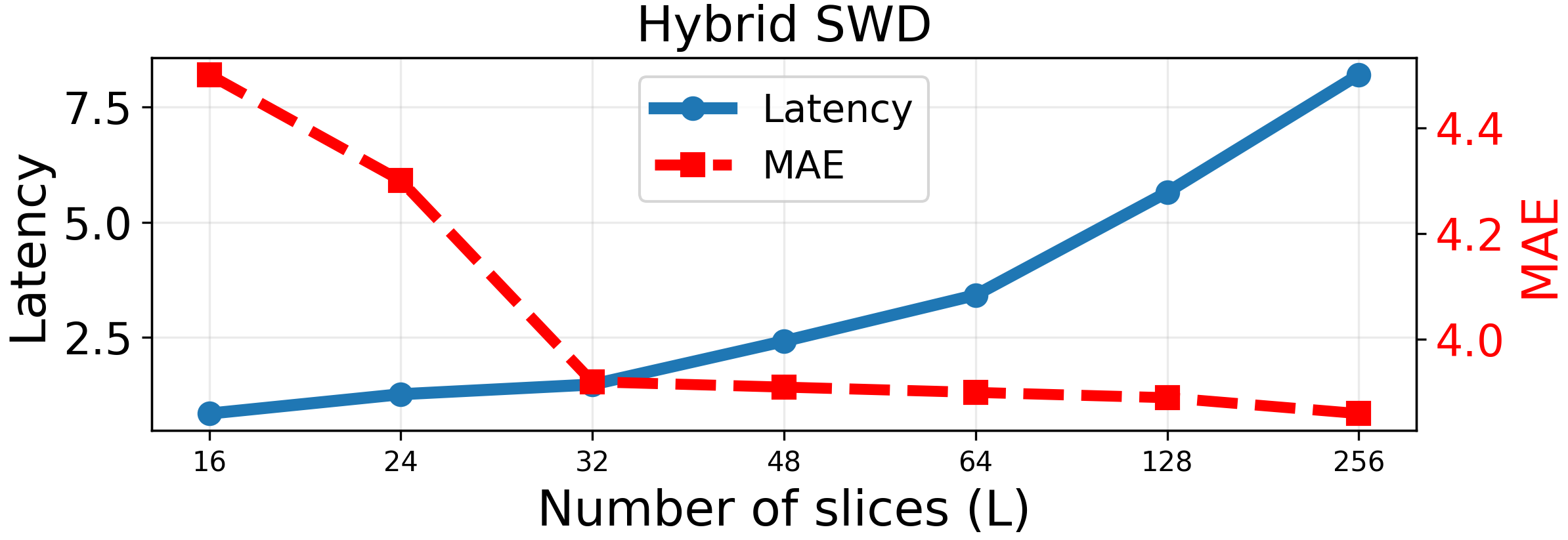}
  \caption{\textbf{Latency + MAE vs. number of slices $L$.} Latency scales linearly with $L$, while MAE stabilizes beyond $L{\approx}32$. Hybrid SWD (red points) achieves low MAE with reduced $L$, optimizing efficiency without sacrificing fidelity.}
  \label{fig:swd_latency_mae}
  \vspace{-1em}
\end{figure}

\sstitle{Scalability}
We evaluate FusionSQL's scalability by varying the number of SWD slices and measuring inference latency (seconds), memory usage, and accuracy. \autoref{fig:swd_latency_mae} shows near-linear latency growth with $L$ while MAE rapidly plateaus, so we can reduce $L$ without sacrificing estimator quality, which is critical for scaling to large workloads and frequent evaluation cycles. \autoref{tab:hybrid_swd_efficiency} compares all-random SWD (AR, $L{=}64$) with Hybrid SWD (HSWD), where configurations $(k,R)$ denote $k$ data-aware PCA directions and $R$ random slices as defined in \autoref{sec:swd_design}. HSWD consistently reduces latency and memory: for instance, HSWD (8,16) cuts latency from 3.42s to 1.27s and memory from 8.9GB to 3.2GB, while maintaining a minor MAE increase of only +0.19\%. Across all settings, HSWD delivers comparable or improved MAE with substantially lower latency and memory, keeping deviations within ${\pm}0.6\%$. This makes FusionSQL practical for large datasets, higher throughput, and routine pre-deployment checks. Therefore, we adopt HSWD in all experiments in \autoref{sec:experiments}.

\begin{table}[h]
\centering
\setlength{\tabcolsep}{4pt}
\caption{\textbf{Efficiency of SWD variants.} Comparison of all-random SWD (AR, $L{=}64$) and hybrid SWD (HSWD). Latency, memory, 95\% CI, and $\Delta$MAE (\%) relative to AR (negative values = improved accuracy) are reported.}
\label{tab:hybrid_swd_efficiency}
\resizebox{0.8\linewidth}{!}{
\begin{tabular}{@{}lccc@{}}
\toprule
\textbf{Config.} & \textbf{Lat. (s)} & \textbf{Mem. (GB)} & \textbf{$\Delta$MAE (\%) [CI]} \\
\midrule
AR-SWD (64) & 3.42 & 8.9 & $0.00{\pm}0.12$ \\
HSWD (4,12) & 0.96 & 2.8 & $+0.60{\pm}0.15$ \\
HSWD (6,16) & 1.12 & 3.0 & $+0.28{\pm}0.14$ \\
HSWD (8,16) & \textbf{1.27} & \textbf{3.2} & $+0.19{\pm}0.13$ \\
HSWD (10,24) & 1.48 & 3.6 & $\mathbf{-0.02}{\pm}0.12$ \\
HSWD (12,32) & 1.94 & 4.5 & $\mathbf{-0.05}{\pm}0.11$ \\
\bottomrule
\end{tabular}
}
\vspace{-1em}
\end{table}

\begin{table}[h]
\centering
\footnotesize
\setlength{\tabcolsep}{4.0pt}
\caption{MAE ($\downarrow$) of classic models (e.g., ATHENA++) across datasets. Each cell reports mean $\pm$ 95\% CI (percentage points). Best in \textbf{bold}; second-best \underline{underlined}.}
\label{tab:mae-classic}
\resizebox{0.9\linewidth}{!}{
\begin{tabular}{lcccc}
\toprule
\textbf{Dataset / Method} &
\textit{ATHENA} &
\textit{ATHENA++} &
\textit{SQLizer} &
\textit{Avg.} \\

\midrule
\multicolumn{5}{l}{\textit{Spider}} \\
BugJudge \cite{liu2025nl2sqlbugs} & 12.0 $\pm$ 1.0 & 11.8 $\pm$ 0.9 & 12.1 $\pm$ 1.0 & 12.0 $\pm$ 1.0 \\
ArenaCmp \cite{zheng2023llmasjudge} &
\underline{10.8 $\pm$ 0.9} &
\underline{10.6 $\pm$ 0.8} &
\underline{10.9 $\pm$ 0.9} &
\underline{10.8 $\pm$ 0.9} \\
FusionSQL-TI & 14.3 $\pm$ 1.1 & 14.1 $\pm$ 1.1 & 14.4 $\pm$ 1.2 & 14.3 $\pm$ 1.2 \\
FusionSQL-LLM & 12.8 $\pm$ 1.1 & 12.6 $\pm$ 1.0 & 12.9 $\pm$ 1.1 & 12.8 $\pm$ 1.1 \\
\textbf{FusionSQL} & \textbf{8.3 $\pm$ 0.6} & \textbf{8.2 $\pm$ 0.7} & \textbf{8.4 $\pm$ 0.8} & \textbf{8.3 $\pm$ 0.7} \\

\midrule
\multicolumn{5}{l}{\textit{Spider~2.0}} \\
BugJudge \cite{liu2025nl2sqlbugs} & 12.8 $\pm$ 1.1 & 12.6 $\pm$ 1.0 & 12.9 $\pm$ 1.1 & 12.8 $\pm$ 1.1 \\
ArenaCmp \cite{zheng2023llmasjudge} &
\underline{11.6 $\pm$ 0.8} &
\underline{11.4 $\pm$ 0.9} &
\underline{11.7 $\pm$ 1.0} &
\underline{11.6 $\pm$ 0.9} \\
FusionSQL-TI & 15.1 $\pm$ 1.1 & 14.9 $\pm$ 1.2 & 15.2 $\pm$ 1.3 & 15.1 $\pm$ 1.2 \\
FusionSQL-LLM & 13.6 $\pm$ 1.0 & 13.4 $\pm$ 1.1 & 13.7 $\pm$ 1.2 & 13.6 $\pm$ 1.1 \\
\textbf{FusionSQL} & \textbf{9.0 $\pm$ 0.6} & \textbf{8.9 $\pm$ 0.7} & \textbf{9.1 $\pm$ 0.8} & \textbf{9.0 $\pm$ 0.7} \\

\midrule
\multicolumn{5}{l}{\textit{SynSQL-2.5M}} \\
BugJudge \cite{liu2025nl2sqlbugs} & 13.0 $\pm$ 1.1 & 12.8 $\pm$ 1.0 & 13.1 $\pm$ 1.1 & 13.0 $\pm$ 1.1 \\
ArenaCmp \cite{zheng2023llmasjudge} &
\underline{11.8 $\pm$ 0.8} &
\underline{11.6 $\pm$ 0.9} &
\underline{11.9 $\pm$ 1.0} &
\underline{11.8 $\pm$ 0.9} \\
FusionSQL-TI & 15.3 $\pm$ 1.2 & 15.1 $\pm$ 1.2 & 15.4 $\pm$ 1.3 & 15.3 $\pm$ 1.3 \\
FusionSQL-LLM & 13.7 $\pm$ 1.1 & 13.5 $\pm$ 1.1 & 13.8 $\pm$ 1.2 & 13.7 $\pm$ 1.1 \\
\textbf{FusionSQL} & \textbf{9.1 $\pm$ 0.6} & \textbf{9.0 $\pm$ 0.7} & \textbf{9.2 $\pm$ 0.8} & \textbf{9.1 $\pm$ 0.7} \\

\midrule
\multicolumn{5}{l}{\textit{CoSQL}} \\
BugJudge \cite{liu2025nl2sqlbugs} & 11.5 $\pm$ 0.8 & 11.3 $\pm$ 0.9 & 11.6 $\pm$ 1.0 & 11.5 $\pm$ 0.9 \\
ArenaCmp \cite{zheng2023llmasjudge} &
\underline{10.2 $\pm$ 0.8} &
\underline{10.0 $\pm$ 0.7} &
\underline{10.3 $\pm$ 0.8} &
\underline{10.2 $\pm$ 0.8} \\
FusionSQL-TI & 13.8 $\pm$ 1.1 & 13.6 $\pm$ 1.1 & 13.9 $\pm$ 1.2 & 13.8 $\pm$ 1.2 \\
FusionSQL-LLM & 12.3 $\pm$ 1.1 & 12.1 $\pm$ 1.0 & 12.4 $\pm$ 1.1 & 12.3 $\pm$ 1.1 \\
\textbf{FusionSQL} & \textbf{7.9 $\pm$ 0.5} & \textbf{7.8 $\pm$ 0.6} & \textbf{8.0 $\pm$ 0.7} & \textbf{7.9 $\pm$ 0.6} \\

\midrule
\multicolumn{5}{l}{\textit{BIRD}} \\
BugJudge \cite{liu2025nl2sqlbugs} & 13.2 $\pm$ 1.1 & 13.0 $\pm$ 1.0 & 13.3 $\pm$ 1.1 & 13.2 $\pm$ 1.1 \\
ArenaCmp \cite{zheng2023llmasjudge} &
\underline{12.0 $\pm$ 0.9} &
\underline{11.8 $\pm$ 0.8} &
\underline{12.1 $\pm$ 0.9} &
\underline{12.0 $\pm$ 0.9} \\
FusionSQL-TI & 15.5 $\pm$ 1.3 & 15.3 $\pm$ 1.2 & 15.6 $\pm$ 1.3 & 15.5 $\pm$ 1.3 \\
FusionSQL-LLM & 13.9 $\pm$ 1.2 & 13.7 $\pm$ 1.1 & 14.0 $\pm$ 1.2 & 13.9 $\pm$ 1.2 \\
\textbf{FusionSQL} & \textbf{9.2 $\pm$ 0.6} & \textbf{9.1 $\pm$ 0.7} & \textbf{9.3 $\pm$ 0.8} & \textbf{9.2 $\pm$ 0.7} \\

\bottomrule
\end{tabular}
}
\vspace{-1em}
\end{table}

\begin{table*}[h]
\centering
\small
\caption{\textbf{Embedding shifts.} Qwen2.5-72B-Instruct embeddings ($8192$-D) are reduced to $4$-D by PCA. We report the mean $\mu$, diagonal variance $\mathrm{diag}(\Sigma)$, and shift vector $\mathbf{s}{=}[{\textit{SD}_F},{\textit{SD}_M},{\textit{SD}_{SW}},\|\mu_A-\mu_B\|_2]^\top$.}
\label{tab:sd_transfers_1_3}
\resizebox{0.8\textwidth}{!}{%
\begin{tabular}{l|c|c|c}
\toprule
 &
A &
B &
\textbf{$\mathbf{s}(A\!\rightarrow\!B)$} \\
\midrule
\parbox[c]{0.25\linewidth}{
A: Q = ``What is the population of Paris?'' (\textit{WikiSQL style})\\
Sch = \texttt{City(City, Country, Population)}\\
B: Q = ``List singer names for concerts in 2014.'' (\textit{Spider style})\\
Sch = \texttt{singer(Singer\_ID, Name), concert(Concert\_ID, Year),}\\
\texttt{singer\_in\_concert(\\
Singer\_ID, Concert\_ID)}} &
$\mu_A=\begin{bmatrix}0.03\\-0.01\\0.00\\0.02\end{bmatrix},\ 
\mathrm{diag}(\Sigma_A)=\begin{bmatrix}0.35\\0.10\\0.06\\0.05\end{bmatrix}$ &
$\mu_B=\begin{bmatrix}0.12\\0.18\\0.05\\0.04\end{bmatrix},\ 
\mathrm{diag}(\Sigma_B)=\begin{bmatrix}1.80\\0.42\\0.30\\0.22\end{bmatrix}$ &
$\begin{bmatrix}
1.95\\
2.60\\
0.90\\
0.22
\end{bmatrix}$ \\
\midrule
\parbox[c]{0.25\linewidth}{
A: Q = ``For each department,\\
return the average salary.''\\
Sch = \texttt{emp(Emp\_ID, Dept\_ID, Salary)}\\
B: SQL = \texttt{SELECT Dept\_ID, Salary FROM emp;}\\
(\textit{missing AVG and GROUP BY})} &
$\mu_A=\begin{bmatrix}0.24\\0.08\\-0.02\\0.04\end{bmatrix},\ 
\mathrm{diag}(\Sigma_A)=\begin{bmatrix}1.10\\0.32\\0.14\\0.11\end{bmatrix}$ &
$\mu_B=\begin{bmatrix}0.03\\0.22\\0.00\\0.02\end{bmatrix},\ 
\mathrm{diag}(\Sigma_B)=\begin{bmatrix}0.62\\0.40\\0.16\\0.12\end{bmatrix}$ &
$\begin{bmatrix}
2.10\\
3.85\\
1.05\\
0.25
\end{bmatrix}$ \\
\midrule
\textbf{Spider $\rightarrow$ BIRD} &
$\mu_A=\begin{bmatrix}0.12\\0.02\\-0.05\\0.01\end{bmatrix},\ 
\mathrm{diag}(\Sigma_A)=\begin{bmatrix}4.90\\0.95\\0.62\\0.41\end{bmatrix}$ &
$\mu_B=\begin{bmatrix}0.05\\0.34\\-0.03\\0.06\end{bmatrix},\ 
\mathrm{diag}(\Sigma_B)=\begin{bmatrix}4.65\\2.70\\0.76\\0.52\end{bmatrix}$ &
$\begin{bmatrix}
0.88\\
3.15\\
0.31\\
0.33
\end{bmatrix}$ \\
\midrule
\textbf{BIRD $\rightarrow$ WikiSQL} &
$\mu_A=\begin{bmatrix}0.06\\0.31\\-0.02\\0.05\end{bmatrix},\ 
\mathrm{diag}(\Sigma_A)=\begin{bmatrix}4.70\\2.60\\0.74\\0.50\end{bmatrix}$ &
$\mu_B=\begin{bmatrix}-0.02\\0.04\\0.01\\-0.02\end{bmatrix},\ 
\mathrm{diag}(\Sigma_B)=\begin{bmatrix}1.05\\0.20\\0.14\\0.11\end{bmatrix}$ &
$\begin{bmatrix}
2.10\\
2.40\\
1.70\\
0.29
\end{bmatrix}$ \\
\midrule
\textbf{WikiSQL $\rightarrow$ Spider} &
$\mu_A=\begin{bmatrix}-0.01\\0.02\\0.00\\-0.01\end{bmatrix},\ 
\mathrm{diag}(\Sigma_A)=\begin{bmatrix}1.10\\0.22\\0.15\\0.10\end{bmatrix}$ &
$\mu_B=\begin{bmatrix}-0.06\\0.30\\0.12\\0.08\end{bmatrix},\ 
\mathrm{diag}(\Sigma_B)=\begin{bmatrix}8.05\\0.86\\0.95\\0.72\end{bmatrix}$ &
$\begin{bmatrix}
4.92\\
5.41\\
1.25\\
0.32
\end{bmatrix}$ \\
\midrule
\textbf{SParC $\rightarrow$ CoSQL} &
$\mu_A=\begin{bmatrix}0.00\\0.01\\0.02\\-0.01\end{bmatrix},\ 
\mathrm{diag}(\Sigma_A)=\begin{bmatrix}2.35\\0.58\\0.44\\0.30\end{bmatrix}$ &
$\mu_B=\begin{bmatrix}-0.08\\0.41\\0.05\\0.12\end{bmatrix},\ 
\mathrm{diag}(\Sigma_B)=\begin{bmatrix}2.40\\0.62\\0.55\\0.39\end{bmatrix}$ &
$\begin{bmatrix}
0.79\\
6.05\\
0.27\\
0.43
\end{bmatrix}$ \\
\bottomrule
\end{tabular}
}
\vspace{-1em}
\end{table*}

\sstitle{Non-Neural Models}
This work primarily targets deep-learning-based Text2SQL systems, which dominate current benchmarks and deliver the strongest empirical performance. To show that FusionSQL is not tied to neural architectures, we also evaluate it on classic non-neural Text2SQL systems, which map a question--schema pair $x=(q,S)$ to a SQL query $y$ via lexicon matching, ontology reasoning, or program synthesis. Without neural representations or labeled shift pairs, we represent each instance as an \textit{input--output} pair $\langle x,y\rangle$ and derive features from their semantic alignment rather than SQL-only cues. FusionSQL is trained on sample sets drawn from FusionDataset and compared against LLM-as-a-judge baselines (BugJudge, ArenaCmp; judge: GPT-5) and two FusionSQL variants: FusionSQL-TI (TF-IDF with cosine similarity), FusionSQL-LLM (last-token pooled embeddings from Qwen2.5-72B). FusionSQL (ours) represents each Text2SQL question--schema pair using BERT-based encodings and characterizes train--test shift through three complementary distributional distance descriptors, namely SD$_F$, SD$_M$, and SD$_{SW}$. As shown in \autoref{tab:mae-classic} on ATHENA \cite{saha2016athena}, ATHENA++ \cite{sen2020athenapp}, and SQLizer \cite{yaghmazadeh2017sqlizer}, FusionSQL achieves the lowest MAE ($\approx$7.5--9.2), with errors increasing with dataset complexity. TF-IDF performs worst due to limited structural awareness, whereas LLM-judge methods miss systematic symbolic errors \cite{oh-etal-2022-dont,gao-etal-2021-simcse,jiang-etal-2022-promptbert}. In comparison, the attention-based FusionSQL-LLM derives embeddings that are highly sensitive to prompt formatting and SQL surface form, whereas the distance-based FusionSQL is more robust to semantic mismatches such as missing joins or incorrect aggregation. Further analysis of \textit{SD}$_F$, \textit{SD}$_M$, and \textit{SD}$_{SW}$ under Text2SQL shifts is provided in \autoref{tab:sd_transfers_1_3}.

\sstitle{Descriptors under Text2SQL Shifts}
As in \autoref{tab:sd_transfers_1_3}, we analyze $(\textit{SD}_F,\textit{SD}_M,\textit{SD}_{SW})$ on two illustrative cases and four transfers using Qwen2.5-72B-Instruct embeddings (mean-pooled last layer, $n{=}450$, PCA$_4$). In the first example, adding joins expands variance and produces moderate global and tail drift ($\textit{SD}_F{=}1.95$, $\textit{SD}_M{=}2.60$) with noticeable shape change ($\textit{SD}_{SW}{=}0.90$). A second example isolates a semantic error (missing \texttt{AVG}/\texttt{GROUP BY}), yielding elevated global shift ($\textit{SD}_F{=}2.10$), large tail deviation ($\textit{SD}_M{=}3.85$), and additional shape loss ($\textit{SD}_{SW}{=}1.05$). At scale, WikiSQL$\rightarrow$Spider shows the strongest drift when introducing nesting and joins ($\textit{SD}_F{=}4.92$, $\textit{SD}_M{=}5.41$). Across all cases, the Euclidean mean shift remains small, showing that it fails to reflect substantial semantic and structural changes. In contrast, the three descriptors capture complementary drifts, yielding an interpretable view of dataset shift. Although SQL errors are discrete, they stem from systematic distribution-level changes in internal representations, motivating FusionSQL's use of higher-order shift descriptors for accuracy estimation.

\vspace{-.5em}
\section{Conclusion \& Future Work}
\label{sec:conclusion}

\vspace{-.5em}
The growing scale of enterprise data demands reliable evaluation of Text2SQL systems on unseen, unlabeled workloads where gold SQL is unavailable or costly. FusionSQL is, to our knowledge, the first framework to enable dataset-level evaluation under these conditions by combining a model-agnostic evaluator, a large-scale realistic FusionDataset, and compact shift descriptors capturing global drift, tail risk, and distributional shape. Extensive experiments show that FusionSQL delivers well-calibrated estimates with low latency, generalizes to unseen models via meta-learning, and remains effective for non-neural systems. Looking ahead, FusionSQL can serve as a lightweight component in automated Text2SQL pipelines and may evolve into a universal, label-free evaluator for unseen Text2SQL models across modalities.

 \section*{Acknowledgement}
The Australian Research Council partially supports this work under the streams of the Discovery Project (Grant No. DP240101108), and the Linkage Project (Grant No. LP240200546).

\section*{AI-Generated Content Acknowledgement}

ChatGPT was used to only improve the language and grammar. The authors reviewed and edited the content and take full responsibility for the content of the publication.

\bibliographystyle{IEEEtran}
\bibliography{IEEEabrv,ref}

\end{document}